\documentclass[runningheads]{llncs}

\usepackage{eccv} 
\usepackage{pdfpages}
\usepackage{subcaption}
\usepackage{wrapfig}

\usepackage{eccvabbrv}
\usepackage{diagbox} 
\usepackage{multirow} 

\usepackage{graphicx}
\usepackage{booktabs}

\usepackage[accsupp]{axessibility}  

\usepackage{hyperref}
\usepackage{comment}
\usepackage[capitalize]{cleveref}

\crefname{figure}{Fig.}{Figs.}
\Crefname{figure}{Figure}{Figures}
\crefname{table}{Table}{Tables}
\crefname{equation}{Eq.}{Eqs.}
\crefname{section}{Sec.}{Secs.}

\usepackage{orcidlink}
\usepackage[colorinlistoftodos]{todonotes}
\newcommand{\langzhe}[1]{\todo[inline,color=green!40]{Langzhe: #1}}
\newcommand{\arxiv}[1]{\todo[inline,color=orange!40]{arxiv: #1}}
\newcommand{\mohamad}[1]{\todo[inline,color=red!40]{Mohamad: #1}}
\newcommand{\joe}[1]{\todo[inline,color=blue!20]{Joe: #1}}
\newcommand{\wenzhen}[1]{\todo[inline,color=gray!20]{Wenzhen: #1}}

\renewcommand{\langzhe}[1]{}
\renewcommand{\arxiv}[1]{}
\renewcommand{\mohamad}[1]{}
\renewcommand{\joe}[1]{}
\renewcommand{\wenzhen}[1]{}

\begin{document}

\title{TouchAnything: Diffusion-Guided 3D Reconstruction from Sparse Robot Touches}
\titlerunning{TouchAnything}

\author{
Langzhe Gu\inst{1,3} \and 
Hung-Jui Huang\inst{2}$^{*}$ \and 
Mohamad Qadri\inst{2}$^{*}$ \and 
Michael Kaess\inst{2} \and 
Wenzhen Yuan\inst{3}
}
\authorrunning{L.~Gu et al.}
\institute{
Tsinghua University \and 
Carnegie Mellon University \and 
University of Illinois Urbana-Champaign \\
\email{$^{*}$Equal contribution}
}

\maketitle

\begin{abstract}
Accurate object geometry estimation is essential for many downstream tasks, including robotic manipulation and physical interaction. Although vision is the dominant modality for shape perception, it becomes unreliable under occlusions or challenging lighting conditions. In such scenarios, tactile sensing provides direct geometric information through physical contact. However, reconstructing global 3D geometry from sparse local touches alone is fundamentally underconstrained. We present TouchAnything, a framework that leverages a pretrained large-scale 2D vision diffusion model as a semantic and geometric prior for 3D reconstruction from sparse tactile measurements. Unlike prior work that trains category-specific reconstruction networks or learns diffusion models directly from tactile data, we transfer the geometric knowledge encoded in pretrained visual diffusion models to the tactile domain. 
Given sparse contact constraints and a coarse class-level description of the object, we formulate reconstruction as an optimization problem that enforces tactile consistency while guiding solutions toward shapes consistent with the diffusion prior. Our method reconstructs accurate geometries from only a few touches, outperforms existing baselines, and enables open-world 3D reconstruction of previously unseen object instances. Our project page is \url{https://grange007.github.io/touchanything/}.

\end{abstract}

\begin{figure}[h]
    \centering
    \includegraphics[width=0.99\linewidth]{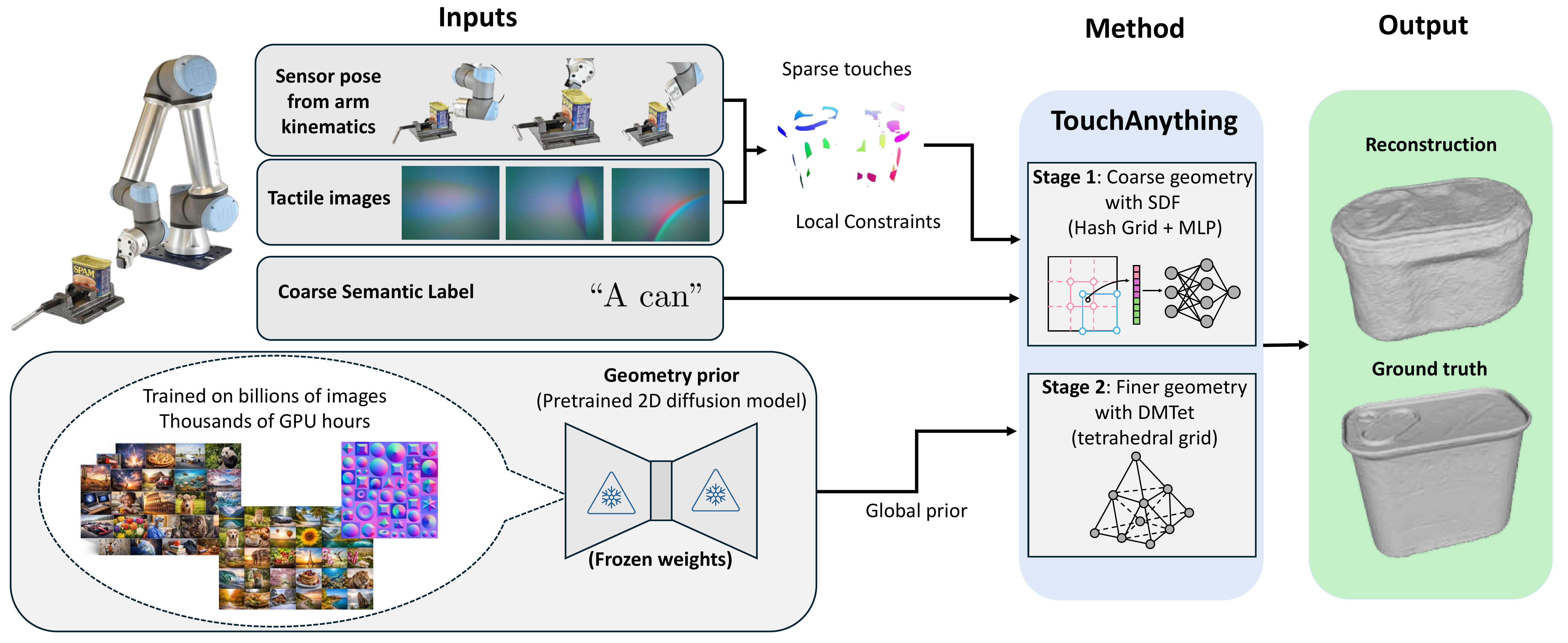}
    \caption{Overview of TouchAnything. Sparse tactile measurements and a class-level text description are combined with a pretrained 2D diffusion model serving as a geometric prior. Local tactile constraints and global diffusion geometric guidance jointly optimize a two-stage geometric representation.}
    \label{fig:overview}
    \wenzhen{text too small. "Reconstruction"->"Reconstructed Shape"}
\end{figure}
\section{Introduction}

Tactile feedback is a fundamental sensory signal that enables humans to interact effectively with the physical world. Touch provides rich information about the geometry and rigidity of objects, from which the nature of possible interactions (e.g., grasps, handling) can be inferred. This becomes especially important under heavy occlusions or challenging lighting conditions, where visual perception may fail. In such situations, the ability to rely on touch alone becomes essential for estimating object geometry and enabling downstream manipulation tasks.

However, geometry estimation from touch alone is inherently underconstrained due to the sparsity of contact measurements. What makes this possible in humans is the presence of strong prior knowledge about object shape and structure.
For example, imagine that you are tasked with finding a pen inside a backpack without visual feedback. You already possess a semantic understanding of what a “pen” typically looks like (e.g. its approximate shape and structure). As you make contact with objects inside the bag, sparse tactile cues are combined with this prior knowledge to refine your belief about the object being touched. Touch does not operate in isolation; rather, it conditions and disambiguates an existing internal model of object geometry.

This raises a natural question: can we equip robots with a similar capability? Specifically, given a robot arm equipped with tactile sensing, can we infer object geometry from sparse touches when guided by a coarse semantic prior and can this be achieved in an open-world setting? \footnote{By open-world, we mean reconstruction of previously unseen object instances without object-specific training.}

Recent work has demonstrated that diffusion models can serve as powerful priors for geometric reasoning. Diffusion-based approaches have been successfully applied to 3D reconstruction, text-to-3D generation, and multi-view consistent shape synthesis, enabling plausible 3D geometry inference even under limited observations or severely constrained settings. Importantly, these models are trained on large-scale visual datasets and have not been specialized to tactile signals. Nevertheless, they implicitly encode strong geometric information that may transfer across sensing modalities. To the best of our knowledge, this work is the first to adapt a general-purpose off-the-shelf pretrained 2D vision diffusion model as a geometric prior for 3D reconstruction from sparse tactile touches.

In contrast to prior touch-based reconstruction methods, which train models on class-specific datasets spanning just a handful of object categories\cite{comi2024touchsdf, wang2025touch2shape, zhang2025end}, we develop a system that leverages the geometric priors encoded in large-scale visual diffusion models trained on billions of internet images to guide 3D reconstruction given sparse tactile measurements. We argue that transferring large-scale visual priors to the tactile domain represents an important step toward open-world generalization, especially in regimes where tactile data is scarce and training specialized generative models is impractical.

Our contributions are as follows:

\begin{itemize}
    \item We introduce \textbf{TouchAnything}, a framework for reconstructing global 3D object geometry from sparse tactile contacts and a coarse semantic prior, enabling open-world 3D reconstruction inference from limited physical interaction.
    
    \item We demonstrate that large-scale pretrained 2D vision diffusion models can be repurposed as geometric priors for tactile reconstruction, transferring visual generative knowledge to the tactile domain without task-specific diffusion training.
    
    \item We provide extensive validation in simulation and real-world robotic experiments, including a  study of reconstruction accuracy under varying numbers of touches and prompt designs.
\end{itemize}

\label{sec:intro}
\section{Related Work}
\subsection{3D Reconstruction from Sparse Observations}
Neural implicit and differentiable geometric representations such as neural radiance fields (NeRF) \cite{mildenhall2021nerf}, signed distance fields (SDFs) \cite{park2019deepsdf}, hybrid mesh-based approaches like DMTet \cite{shen2021deep}, and more recently 3D Gaussian Splatting \cite{kerbl20233d}, have become dominant frameworks for 3D reconstruction. These methods introduce differentiable parameterizations of geometry and appearance, making them well-suited for end-to-end optimization. Beyond RGB-based reconstruction, their flexibility has enabled applications across diverse sensing modalities \cite{qadri2023neural, qadri2024aoneus, qu2024z, lin2025acoustic, rafidashti2025neuradar, kung2025radarsplat, zhao2024tclc}, including tactile sensing \cite{comi2024touchsdf, suresh2024neuralfeels}. However, regardless of the underlying representation or sensing modality, 3D reconstruction becomes severely underconstrained when observations provide only limited geometric coverage of the underlying surface \cite{niemeyer2022regnerf, wang2023sparsenerf}. In few-view settings, implicit and differentiable models alone are insufficient to solve for global geometry without additional regularization. This challenge is further amplified in tactile reconstruction, where measurements come from small local contact patches rather than partial global image observations. 

\subsection{Diffusion Models as Geometric Priors}
Recent advances in diffusion modeling have demonstrated that large-scale generative models implicitly encode rich knowledge about object geometry. DreamFusion \cite{poole2022dreamfusion} introduced the use of pretrained 2D text-to-image diffusion models to optimize 3D representations via score distillation sampling (SDS), enabling text-to-3D generation without direct 3D supervision. Subsequent works \cite{lin2023magic3d, wang2023prolificdreamer, qiu2024richdreamer}, improved fidelity and enhanced geometric detail in diffusion-guided 3D synthesis. Fantasia3D \cite{chen2023fantasia3d} and RichDreamer \cite{qiu2024richdreamer} explicitly incorporate geometric signals such as surface normals and depth to better disentangle shape and appearance during diffusion-guided 3D generation. These works demonstrate that diffusion models can effectively leverage explicit geometric cues to improve the quality of the generated geometry. In our setting, we similarly render normal maps from the evolving 3D geometry. However, unlike prior methods, we additionally enforce consistency with real local surface normal measurements obtained from image-based tactile sensors. We therefore combine global diffusion guidance with physical local constraints imposed by robot contact. Tactile DreamFusion \cite{gao2024tactile} similarly incorporates tactile signals into text-to-3D synthesis, but its primary objective remains asset generation, whereas our goal is tactile-conditioned 3D reconstruction constrained by real tactile measurements. 

Beyond text-to-3D synthesis, sparse 3D reconstruction has been addressed by learning task-specific shape completion priors from curated 3D datasets \cite{chibane2020implicit, yuan2018pcn}. More recently, diffusion-based completion models \cite{kasten2023point, schaefer2024sc} train 3D diffusion priors to regularize underconstrained geometric inference and recover plausible structure from partial observations. Unlike these approaches, which require supervised training on 3D shape collections, we leverage a pretrained 2D diffusion model as a transferable geometric prior without task-specific diffusion training.

Several recent works \cite{wu2024reconfusion, zou2024sparse3d, Ni_2025_CVPR} also employ diffusion models as geometric priors for sparse-view 3D reconstruction from RGB images, where diffusion guidance regularizes geometry estimated from limited visual coverage. In contrast, our setting relies solely on sparse local tactile contacts, which provide highly localized surface measurements without global image-level constraints. Inferring global geometry from such physical interaction signals is a different and a more severely underconstrained problem.

\subsection{Tactile-based Shape Reconstruction}
Image-based tactile sensors provide useful geometric information (e.g. contact location and surface normals) which have been leveraged for object state estimation and tracking during interaction ~\cite{huang2024normalflow, qadri2022incopt, qadri2024learning, huang2025gelslam} as well as 3D reconstruction.
TouchSDF \cite{comi2024touchsdf} predicts local surface geometry from vision-based tactile sensors and learns an SDF encoding the object shape. More recently, diffusion-based tactile reconstruction methods \cite{wang2025touch2shape, zhang2025end} train conditional diffusion models for tactile-conditioned 3D reconstruction using object-level datasets such as ShapeNet and ABC. These approaches rely on supervised training with ground-truth 3D geometry from fixed object categories (e.g., guitars, bottles), which may limit generalization beyond seen categories and involve computationally intensive task-specific training of diffusion models. Other works \cite{suresh2022shapemap, suresh2024neuralfeels, comi2025snap, swann2024touch, fang2025fusionsense} fuse tactile and visual measurements for 3D reconstruction benefiting from the global visual coverage provided by vision. However, they differ from our setting where touch provides the primary geometric signal. Overall, prior tactile reconstruction methods either depend on dense measurements, multi-modal sensing, or expensive training of task-specific generative priors. To the best of our knowledge, reconstructing global object geometry from sparse tactile contacts using pretrained off-the-shelf 2D diffusion models remains unexplored. 
\section{Methodology}
\begin{figure}[t]
    \centering
    \includegraphics[width=1.0\linewidth]{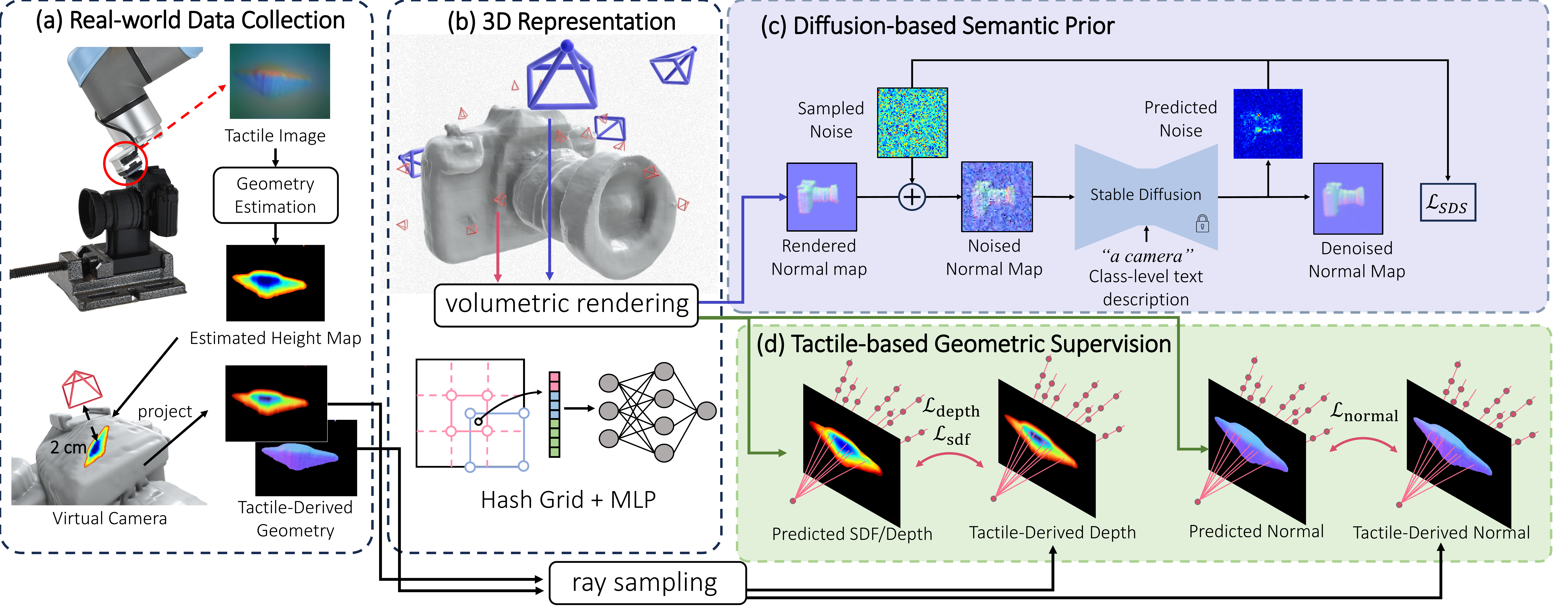}
    \caption{\textbf{TouchAnything reconstruction pipeline.}
We use a GelSight tactile sensor mounted on a robotic arm to collect raw tactile measurements, which are converted into local depth and surface normal maps (\cref{sec:tactile_data_acquisition}). These measurements provide sparse geometric supervision for learning a neural signed distance field (SDF) by enforcing local constraints through depth and normal losses. Normal maps rendered from the evolving 3D geometry are fed into a pretrained Stable Diffusion model, which provides global geometric guidance through score distillation sampling (SDS). By jointly enforcing tactile consistency combined with a pretrained 2D diffusion model and a class-level text description, the system reconstructs globally consistent 3D geometry from sparse contact measurements (\cref{learningcoarsegeometry}).}
    \label{fig:overview}
\end{figure}

Our goal is to reconstruct the 3D geometry of an arbitrary object using only sparse tactile readings and a minimal class-level text description (e.g. ``a camera'', ``a bottle''). We assume that the contact locations of tactile readings are known from robot kinematics. The text description provides only a weak semantic prior at the category level, reflecting realistic scenarios where a robot knows the object class but not the specific geometry, such as searching for a key in the wallet.

To solve this problem, we propose TouchAnything, a diffusion-guided method for 3D reconstruction from sparse tactile inputs. Instead of utilizing a model trained on limited object categories\cite{comi2024touchsdf, zhang2025end}, we leverage Stable Diffusion \cite{Rombach_2022_CVPR}, a general-purpose generative model, to guide reconstruction, conditioned on the weak text description. This design makes our method much more general, enabling reconstruction of arbitrary objects without pre-training or class-specific data collection. In TouchAnything, tactile readings are first converted into local depth maps using established methods \cite{wang2021wedge, suresh2022shapemap}, and represent it as virtual camera observations to ensure compatibility with the vision-centric reconstruction pipeline. We then adopt a coarse-to-fine reconstruction strategy inspired by Magic3D \cite{lin2023magic3d} and optimize the object geometry by jointly enforcing consistency with tactile-derived geometry and alignment with the class-level text description.

\arxiv{fixed class-specific description}

\subsection{Deriving Local Geometry from Tactile Sensing} \label{sec:tactile_data_acquisition}
We use a GelSight sensor \cite{yuan2017gelsight} to collect tactile data. GelSight is a vision-based tactile sensor composed of a soft elastomer sensing surface, an integrated illumination system, and a camera. Upon contact, the elastomer deforms, changing the reflected illumination pattern, which is captured by the camera and used to reconstruct the contact surface geometry via photometric stereo. A sample tactile image is shown in \cref{fig:overview}a. In our work, GelSight images are obtained from both real-world experiments and simulations. From these images, we estimate the local contact geometry and convert it into virtual camera observations for compatibility with the vision-centric reconstruction pipeline.

\subsubsection{Geometry Estimation from Real-World Tactile Data}
Given a tactile image $\mathbf{T}$ collected by a physical GelSight sensor, we adopt the learning-based method in \cite{wang2021wedge} to estimate per-pixel surface gradients from the RGB values and coordinates of each pixel using a three-layer MLP. The predicted gradients are then integrated using a fast Poisson solver\cite{yuan2017gelsight} to recover the surface depth map. The contact mask is obtained by thresholding the estimated depth map.
\subsubsection{Geometry Estimation from Simulated Tactile Data}

Given a photorealistic simulated GelSight image, we adopt the learning-based method in \cite{suresh2022shapemap}, using a multi-head U-Net \cite{ronneberger2015unet} to jointly predict the depth map and contact mask. The network is trained on 20k simulated GelSight images generated from contacts with 78 YCB objects \cite{ycb2017}. Compared to the real-world scenario, we use a more complex geometry estimation model for simulated tactile data because larger-scale training data are available in simulation.

\subsubsection{Tactile-Derived Geometry as Virtual Camera Observation}

For each tactile reading $\mathbf{T}_i$, we place a virtual camera $\mathbf{C}_i$ $2.0$ cm behind the contact patch and convert the tactile-estimated depth map and contact mask into virtual camera observations consisting of depth and normal maps with an associated contact mask. The reconstruction process then operates on these tactile-derived geometries in the form of virtual camera observations rather than the raw tactile readings, ensuring compatibility with the vision-centric reconstruction pipeline. The full data collection pipeline is shown in \cref{fig:overview}a.

\subsection{Stage 1: Learning Coarse Geometry of the Object}
\label{learningcoarsegeometry}
We model the coarse object geometry using a SDF-based neural implicit representation $f_\theta: \mathbb{R}^3 \rightarrow \mathbb{R}$, implemented using a Neuralangelo-style \cite{li2023neuralangelo} multi-resolution 3D hash-grid features followed by an MLP (\cref{fig:overview}b). The function $f_\theta$ maps a 3D coordinate to its truncated signed distance to the object surface, and its zero level set defines the reconstructed shape. We optimize $\theta$ by minimizing a joint objective that enforces geometric consistency with the tactile-derived geometry while encouraging semantic consistency through a diffusion-based prior on the class-level text description.

\subsubsection{Tactile-based Geometric Supervision}
\label{tactilelossstage1}

Consider reconstructing an object that is touched $K$ times at different locations, producing tactile readings  $\mathbf{T}_1, ..., \mathbf{T}_K$. As described in~\cref{sec:tactile_data_acquisition}, we convert each tactile reading into a virtual camera observation consisting of depth maps, normal maps, and contact regions. We use these observations to impose sparse geometric constraints on the object surface represented by the SDF function $f_\theta$. However, to impose these constraints, we need to first represent the virtual camera observations in a ray-based form compatible with our reconstruction algorithm (\cref{fig:overview}d). 

\arxiv{fixed "K" for "k"}

Let $\mathcal{R}$ denote the union of all rays cast from the $K$ virtual cameras $\{\mathbf{C}_K\}$, where each ray originates from a virtual camera center and passes through a pixel within the contacted region. For each ray $r \in \mathcal{R}$, the tactile-derived depth and normal maps in the virtual camera frame provide the depth observation $d(r)$ and surface normal observation $\mathbf{n}(r)$ along that ray.

To enforce geometric constraints from the tactile-derived ray observations $d(r)$ and $\mathbf{n}(r)$, we adopt the volumetric rendering formulation of Neuralangelo~\cite{li2023neuralangelo}. For each ray $r \in \mathcal{R}$, we use the SDF function $f_\theta$ along the ray to compute the predicted depth $d_\theta(r)$ and surface normal $\mathbf{n}_\theta(r)$. These computed quantities are differentiable with respect to the SDF parameters $\theta$, allowing us to enforce geometric consistency by minimizing the discrepancy between the SDF-computed depth and normal values and the tactile-derived depth and normal values:

\begin{equation}\label{eq:depth_and_normal_loss}
\mathcal{L}_{\text{depth}} =
\mathbb{E}_{r \sim \mathcal{R}}
\left[ \left| d(r) - d_{\theta}(r) \right| \right],
\qquad\quad
\mathcal{L}_{\text{normal}} =
\mathbb{E}_{r \sim \mathcal{R}}
\left[ \left\| \mathbf{n}(r) - \mathbf{n}_{\theta}(r) \right\|_{1} \right]
\end{equation}
Beyond supervision on these computed quantities, we incorporate two additional supervision signals following \cite{Azinovic_2022_CVPR}. The first supervises the signed distance values near the observed surface, and the second enforces freespace constraints along the ray up to the surface. To supervise the SDF, for each ray $r$ with a tactile-derived depth observation $d(r)$, we sample depths $s$ within a truncated band $\mathcal{S}^{\mathrm{sdf}}_r = [d(r) - \delta,\; d(r) + \delta]$ around the surface, where $\delta$ denotes the truncation distance. Let $\mathbf{x}(r, s)$ denote the 3D point at depth $s$ along ray $r$, the SDF loss is:

\begin{equation}\label{eq:sdf_loss}
\mathcal{L}_{\text{sdf}}
=
\mathbb{E}_{r \sim \mathcal{R}}
\;
\mathbb{E}_{s \sim \mathcal{S}^{\mathrm{sdf}}_r}
\left|
f_\theta(\mathbf{x}(r, s))
-
\left(s - d(r)\right)
\right|.
\end{equation}
To enforce freespace constraints, we additionally sample points in $\mathcal{S}^{\mathrm{fs}}_r = [0,\; d(r) - \delta]$, which extends from the virtual camera center to a distance $\delta$ before the surface. Since this region should be physically occupied by the tactile sensor and should remain free of objects, we penalize the predicted signed distance to be smaller than the distance $\delta$:

\begin{equation}\label{eq:freespace_loss}
\mathcal{L}_{\mathrm{fs}}
=
\mathbb{E}_{r \sim \mathcal{R}}
\;
\mathbb{E}_{s \sim \mathcal{S}^{\mathrm{fs}}_r}
\left[
\operatorname{ReLU}\!\left(\delta - f_\theta(\mathbf{x}(r,s))\right)^2
\right].
\end{equation}
We optimize a weighted combination of these four losses to enforce geometrical consistency with the tactile observations. In practice, the expected values of the losses are estimated by sampling a batch of rays $r \in \mathcal{R}$ during each training iteration.

\arxiv{added sampling for estimating expection}

\subsubsection{Diffusion-based Prior}
\label{diffusionprior}
To guide the reconstructed geometry to align with the class-level text description (e.g. ``a camera''), we incorporate a diffusion prior. In particular, we use Stable Diffusion \cite{Rombach_2022_CVPR}, a general-purpose generative model, and apply score distillation sampling (SDS) \cite{poole2022dreamfusion} to impose geometric and semantic guidance during optimization (\cref{fig:overview}c).

At each optimization step, we uniformly sample a virtual camera pose from a sphere centered at the object and render a surface normal image $\mathbf{N}_\theta$ via volumetric rendering of the SDF $f_\theta$. Following Fantasia3D \cite{chen2023fantasia3d}, we supervise the geometry using the normal map rather than an RGB image to focus on providing fine surface details. We denote $\mathbf{z}(\mathbf{N}_\theta)$ as the latent feature obtained by passing the rendered normal map $\mathbf{N}_\theta$ through the Stable Diffusion VAE encoder. The SDS gradient with respect to the SDF parameters $\theta$ is:

\begin{equation}\label{eq:sds_loss}
\nabla_\theta \mathcal{L}_{\text{SDS}}
= \mathbb{E}_{t, \epsilon}
\left[
w(t) \big( \hat{\epsilon}_\phi(\mathbf{z}_t(\mathbf{N}_\theta); y, t) - \epsilon \big) \frac{\partial \mathbf{z}(\mathbf{N}_\theta)}{\partial \theta}
\right]
\end{equation}
where $\hat{\epsilon}_\phi$ is the noise predicted by Stable Diffusion, $y$ is the text description, and $\epsilon$ is the actual noise added to the latent $\mathbf{z}$ to produce the noisy version $\mathbf{z}_t$. Additionally, $t$ denotes the diffusion timestep, and $w(t)$ is a timestep-dependent weighting function. By backpropagating this SDS gradient, we refine the SDF parameters $\theta$, so the reconstructed geometry matches the text description $y$.

\subsubsection{Training Schedule}
The final optimization objective combines the tactile-defined geometric losses (\cref{eq:depth_and_normal_loss}, \cref{eq:sdf_loss}, and \cref{eq:freespace_loss}), the diffusion-guided SDS loss (\cref{eq:sds_loss}), and the Eikonal regularization term. We adopt a two-stage training strategy. In the warm-up stage (steps 0–1000), optimization is driven only by tactile supervision to establish a geometry consistent with the tactile readings. In the joint refinement stage (steps 1000-7000), we optimize the model with both tactile supervision and SDS guidance, where the diffusion model operates on $64\times64$ rendered normal images with a batch size of $8$.

\subsection{Stage 2: Learning Fine Geometry of the Object}
To refine geometric details beyond the coarse reconstruction, we further learn fine object geometry using DMTet, an explicit tetrahedral-grid SDF representation. In \cref{learningcoarsegeometry}, the diffusion-based prior is constrained to apply to low-resolution rendered images of $64 \times 64$, because the underlying geometry is represented by an MLP that predicts SDF values and requires expensive per-ray field queries for volumetric rendering. Rendering at higher resolutions is therefore computationally prohibitive, which limits the diffusion prior’s ability to recover fine geometric details. In contrast, DMTet adopts a fully explicit representation. It discretizes the signed distance field onto a tetrahedral grid with vertices $\mathcal{V}=\{v_i\}_{i=1}^{N_v}$ and parameterizes the geometry by per-vertex signed distance values $\{s_i\}_{i=1}^{N_v}$ and per-vertex offsets $\{\Delta v_i\}_{i=1}^{N_v}$, where $N_v$ is the number of grid vertices. This representation eliminates the need for MLP evaluation and per-ray field queries. The object surface can be extracted using differentiable marching tetrahedra and rendered with differentiable rasterization, which is significantly more efficient than rendering with the Neuralangelo-style SDF representation used in \cref{learningcoarsegeometry}. Consequently, DMTet enables rendering at a much higher resolution of $512\times512$ with batch size $4$. 

Let $\phi = \{\{s_i\}, \{\Delta v_i\}\}$ denote the learnable parameters of DMTet. We initialize ${s_i}$ by querying the SDF of the optimized coarse geometry from \cref{learningcoarsegeometry} at each grid vertex. Similar to learning the coarse geometry, we optimize $\phi$ by minimizing the weighted sum of the depth loss $\mathcal{L}_\text{depth}$,  normal loss $\mathcal{L}_\text{normal}$, and the SDS loss $\mathcal{L}_\text{SDS}$. Different from stage 1, these losses are now computed through efficient differentiable rasterization instead of ray casting and MLP evaluation. In particular, the SDS loss is evaluated at a much higher rendering resolution, enabling the diffusion-based prior to recover finer geometric details. Beyond these losses, we additionally include the normal consistency loss~\cite{ravi2020pytorch3d} that penalizes angular deviations between adjacent vertex normals, encouraging locally smooth surface geometry:
\begin{align}
\mathcal{L}_{\text{nc}} = \frac{1}{|\mathcal{E}|} \sum_{(i, j) \in \mathcal{E}} \left( 1 - \cos(\mathbf{n}_i, \mathbf{n}_j) \right),
\end{align}
where $\mathcal{E}$ denotes the set of mesh edges, and $\mathbf{n}_i$ and $\mathbf{n}_j$ are the unit vertex normals at the endpoints of edge $(i,j)$.

\arxiv{removed typo "s"}
\arxiv{Didn't move DMTet image here, because the previous method image was used for the whole Method Section, not just stage 1. Adding a new pipeline image here seems making it imbalanced.}
\arxiv{cite pytorch3d for normal consistency loss. DMTet didn't use this loss.}

\begin{figure}[h]
\centering
\includegraphics[width=0.7\linewidth]{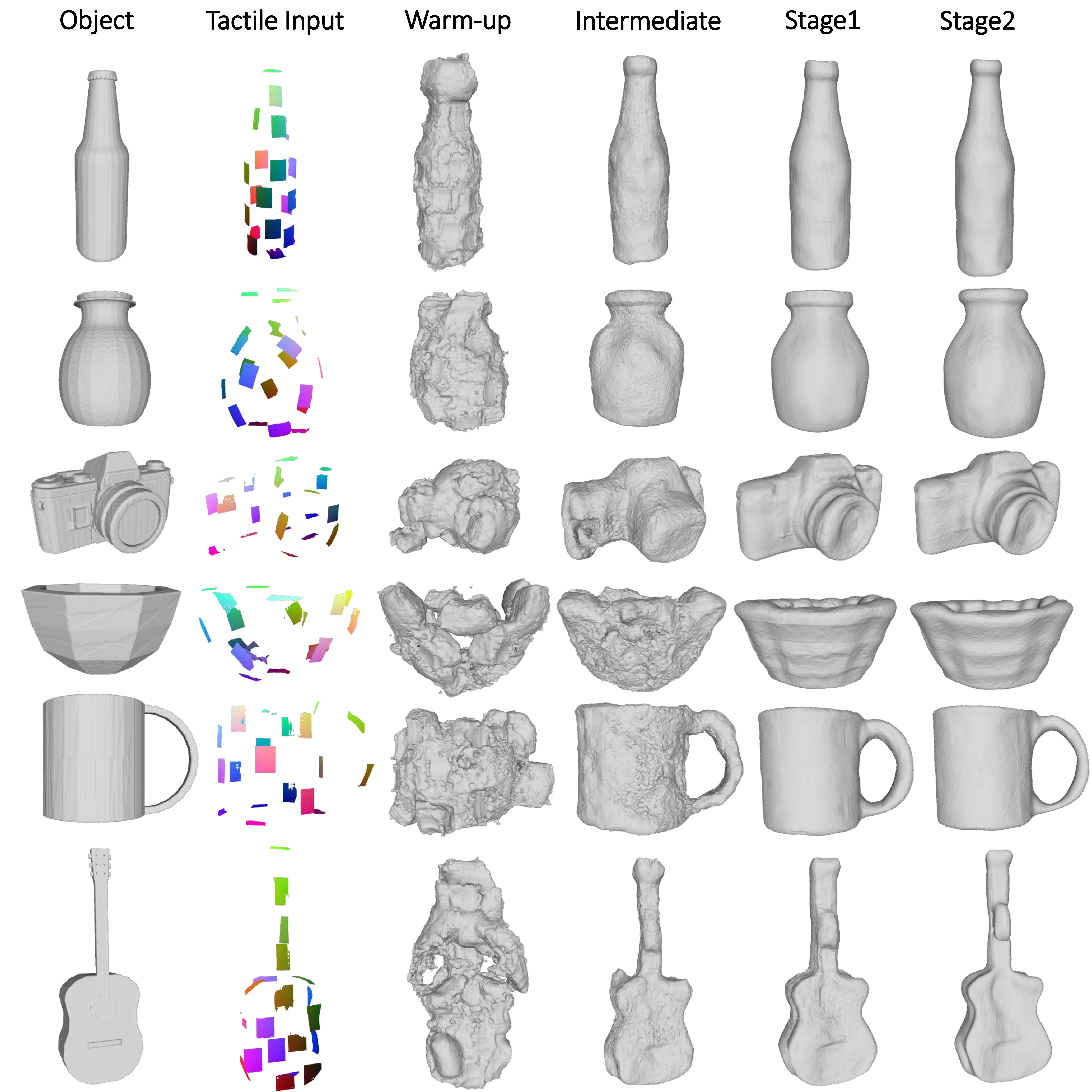}
\caption{Visualization of the simulation results. From left to right, we show the ground truth object, the tactile measurements, the result from the warm-up stage where only tactile observations are used, followed by intermediate, stage 1, and stage 2 reconstructions. The warm-up stage provides a good initialization of the shape before incorporating the diffusion model. We also note that the stage 1 results are close to stage 2 because the simulated objects contain limited geometric detail.}
\label{fig:simulation_visualization}

\end{figure}

\section{Experiments and Results}

We evaluate the reconstruction performance of TouchAnything in simulation and compare it with baseline methods. We additionally demonstrate its qualitative performance in real-world scenarios and conduct ablation studies to examine the impact of different modality inputs. All objects were trained on a single NVIDIA A100 GPU which takes $\sim 1$  hour for stage 1 and 40 minutes for stage 2. 

\subsection{Simulation Experiments}

\subsubsection{Data Collection}
We evaluate TouchAnything on $280$ objects from ShapeNetCore.V2~\cite{shapenet2015}. We use the same object subset as TouchSDF~\cite{comi2024touchsdf}, provided by its authors, to enable direct comparison with it and subsequent work \cite{wang2025touch2shape}. The objects span six categories: bowls, bottles, cameras, jars, guitars, and mugs. The class-level text prompt associated with each category is defined respectively as “a bowl”, “a bottle”, “a camera”, “a jar”, “a guitar”, and “a mug”. For each object, we generate 20 simulated tactile images as input to TouchAnything. This is achieved through a tactile simulation pipeline designed to closely approximate how a robot interacts with objects in real-world scenarios. We use Taxim~\cite{si2022taxim}, an example-based photorealistic tactile simulator, to produce GelSight images. For each object mesh, we uniformly sample contact locations (see Supplementary Material for details), align the contact orientation with the local surface normal, and press the object to a predefined depth. Open3D ray casting~\cite{zhou2018open3d} is then used to compute the resulting contact depth map, which Taxim converts into a photorealistic GelSight image with a resolution of $320 \times 240$ corresponding to a sensing area of $2.0 \text{cm}\times1.5 \text{cm}$.

\arxiv{added "This is achieved through a tactile simulation pipeline designed to closely approximately how a robot interacts with objects in real-world scenarios"}

\subsubsection{Baseline Methods}
We compare our approach to two tactile-based 3D reconstruction methods. \textbf{TouchSDF} \cite{comi2024touchsdf} reconstructs 3D geometry using TacTip tactile sensors and is a common baseline in tactile-based 3D reconstruction research. It represents object geometry using DeepSDF, a neural implicit representation pretrained on a specific dataset, which provides a dataset-specific shape prior during reconstruction. \textbf{Touch2Shape} \cite{wang2025touch2shape} uses an active tactile exploration strategy and additionally trains a diffusion model on tactile data to provide a generative shape prior during reconstruction. In contrast, our approach uses a general-purpose model guided only by minimal text descriptions, enabling open-world reconstruction of arbitrary objects.

\arxiv{fixed class-specific description of TouchSDF}

\begin{table}[!h]
    \centering
    \caption{Quantitative Results on the simulation data. We show the Earth Mover's Distance (EMD) metric for various methods on test objects with 20 robot touches.}
    \resizebox{0.55\linewidth}{!}{
        \begin{tabular}{lccccc}
            \toprule
            Category & TouchSDF \cite{comi2024touchsdf} & Touch2Shape~\cite{wang2025touch2shape} & Ours \\
            \midrule
            Bottle & $0.047 \pm 0.024$ & $0.041$ & $ \mathbf{0.035} \pm 0.020 $ \\
            Bowl   & $0.048 \pm 0.017$ & $0.049$ & $\mathbf{0.039} \pm 0.010 $ \\
            Camera & $0.092 \pm 0.043$ & $0.056$ & $\mathbf{0.047} \pm 0.025 $ \\
            Guitar & $0.155 \pm 0.087$ & $0.064$ & $\mathbf{0.060} \pm 0.025 $ \\
            Jar    & $0.071 \pm 0.038$ & $0.055$ & $ \mathbf{0.053} \pm 0.023 $ \\
            Mug    & $0.066 \pm 0.018$ & $0.049$ & $ \mathbf{0.044} \pm 0.008 $\\
            \bottomrule
        \end{tabular}
    }
    \label{tab:emd_results}
\end{table}

\arxiv{fixed typo in table title: "Earth Moving" to "Earth Mover's"}

\langzhe{"Earth Moving" to "Earth Mover's"}
\subsubsection{Experimental Results} We evaluate the reconstruction results using Earth Mover's Distance (EMD), which is widely used in touch-based reconstruction and better reflects visual quality \cite{comi2024touchsdf}. The result is reported in \cref{tab:emd_results}. TouchAnything achieves better reconstruction performance across all categories compared to both baselines, highlighting our method's ability to leverage the rich geometric priors encoded in off-the-shelf 2D diffusion models. In \cref{fig:simulation_visualization}, we show sample results from the warmup, stage 1 and stage 2.  We note that the warmup stage provides a good initialization of the geometry before further refinement with the diffusion prior while the later incorporation of the diffusion model further improves the reconstruction quality.

\subsection{Real-world Experiments with a Robot}
\begin{figure}[!h]
\centering
\includegraphics[width=0.9\linewidth]{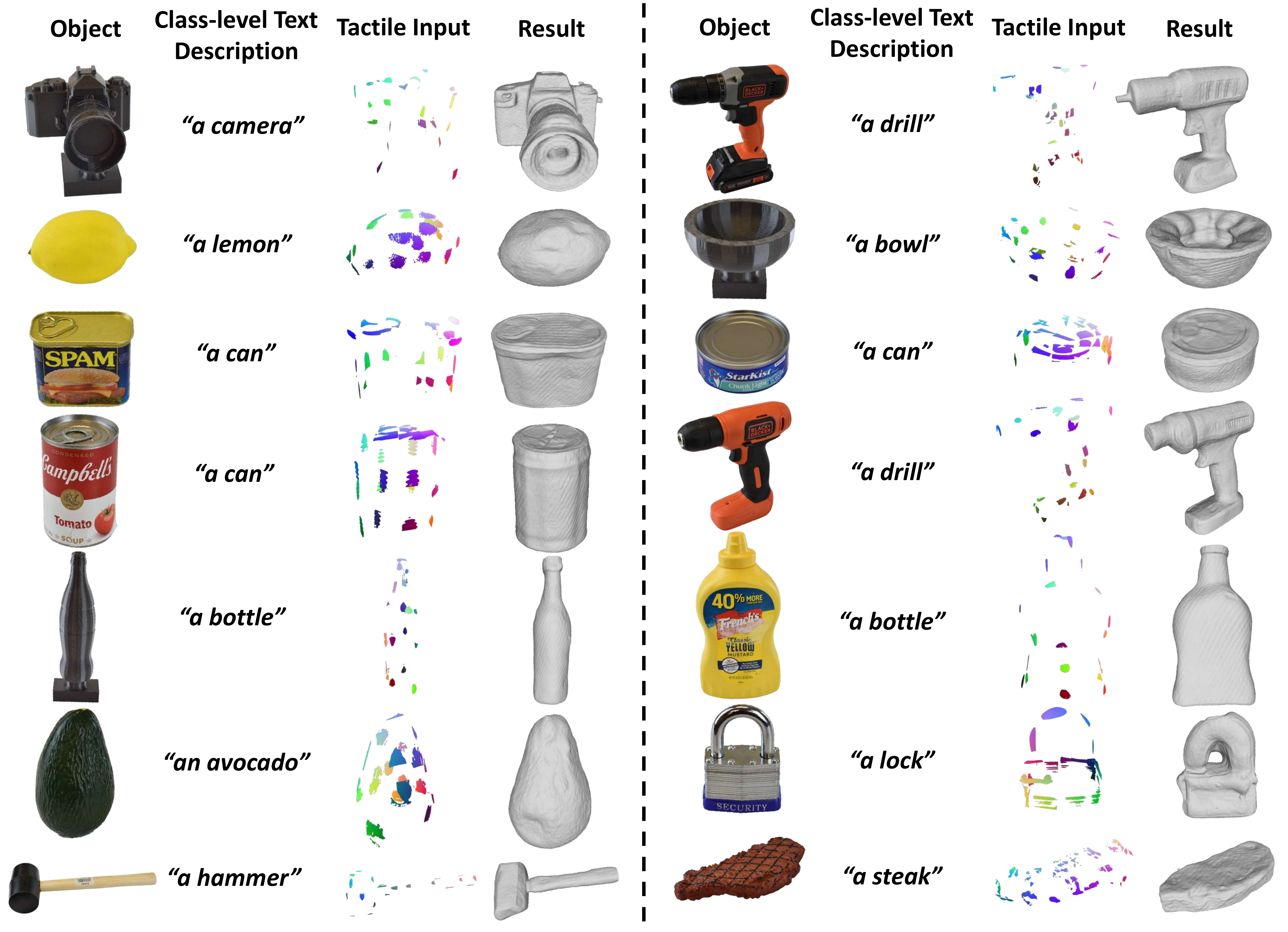}
\caption{Real-world reconstruction results with TouchAnything. For each object, we show the class-level text description, an image of the real object, and the 20 tactile measurements used for reconstruction. The rightmost column shows the results obtained from stage 2.}
\label{fig:real_world_results}
\end{figure}

\arxiv{use brighter objects & change font}

\subsubsection{Hardware} The real-world experiments are conducted with a 6-DOF UR5e robot arm equipped with a Gelsight Mini tactile sensor. The sensor operates at a resolution of $320 \times 240$, corresponding to a sensing area of $2.0 \text{cm}\times1.5 \text{cm}$. For real-world evaluation, we selected 14 objects in total, including 6 objects selected from the YCB dataset, 3 3D-printed objects chosen from ShapeNetCore.V2~\cite{shapenet2015} and 5 household objects. The 3D-printed objects were designed to be mounted on a cuboid base. During the experiments, all objects were rigidly fixed to the table using a dedicated mount.
\langzhe{number of objects may change depending on our results}

\subsubsection{Data Collection}
We collected the real-world data by operating the robot arm to make contact with the object. The contact poses were selected to best cover the surface of the object uniformly. A GelSight image is collected after every touch, and the pose of the sensor is calculated using the forward kinematics of the robot arm.

\subsubsection{Experimental Results}
\cref{fig:real_world_results} presents the tactile input and reconstruction results for all real-world objects. For each object, 20 touches are applied uniformly across the surface. Despite the sparse tactile observations, TouchAnything achieves good reconstruction of the object geometry. For complex objects such as the camera and drill, large portions of the object remain untouched. Nevertheless, our diffusion prior correctly completes these regions with detailed shapes consistent with the semantic description. More importantly, as shown in \cref{fig:real_world_results}, our method successfully reconstructs all tested real-world objects. \begin{wrapfigure}{r}{0.38\textwidth}
    \vspace{-10pt}
    \centering
    \includegraphics[width=0.32\textwidth]{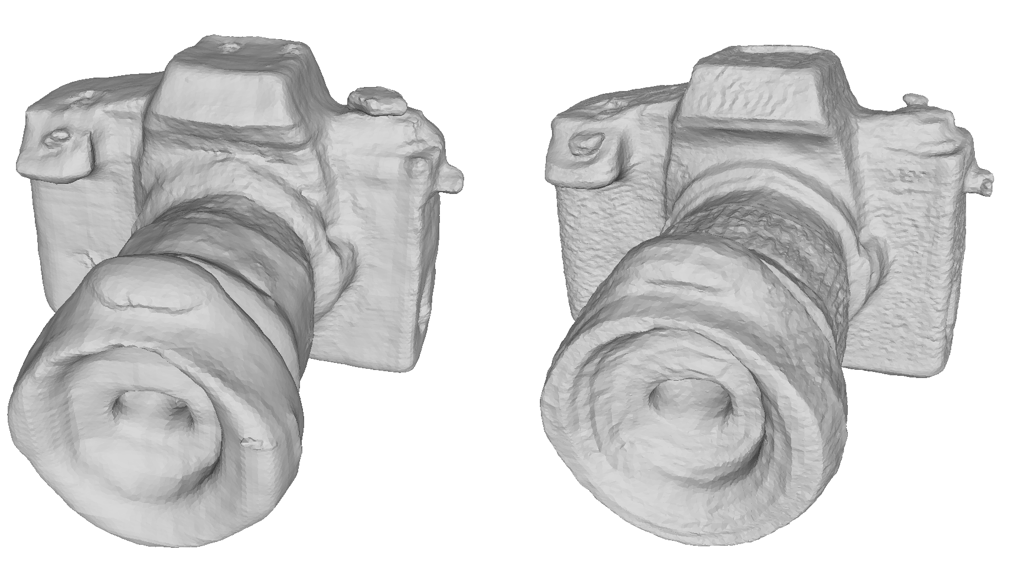}
  \caption{Stage 1 results on the left and stage 2 result on the right. Note the finer details recovered via the refinement step.}
    \label{samplediff}
    \vspace{-10pt}
\end{wrapfigure}This is possible because we use a general-purpose generative model rather than class-specific models. In contrast, prior methods \cite{comi2024touchsdf, zhang2025end, wang2025touch2shape} can typically reconstruct only objects from categories seen during training which highlights the stronger generalization capability of our approach. \cref{samplediff} shows the fine geometric details recovered during the stage 2 refinement step compared to the coarse geometry obtained in stage 1. Additional qualitative results are available in the supplementary material.

\begin{figure}[!h]
\centering
\includegraphics[width=0.78\linewidth]{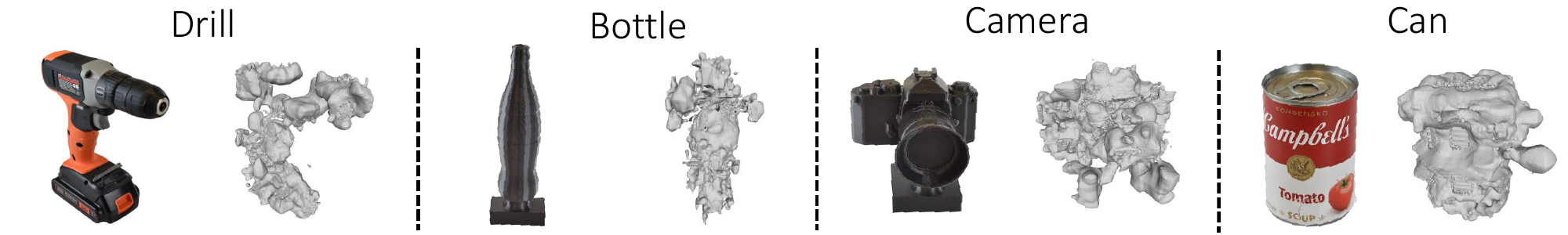}
\caption{Reconstructions using only tactile observations  after removing the diffusion prior. The reconstructions degrade significantly.}
\label{fig:ablationtactonly}
\end{figure}

\begin{figure}[!h]
\centering
\includegraphics[width=0.93\linewidth]{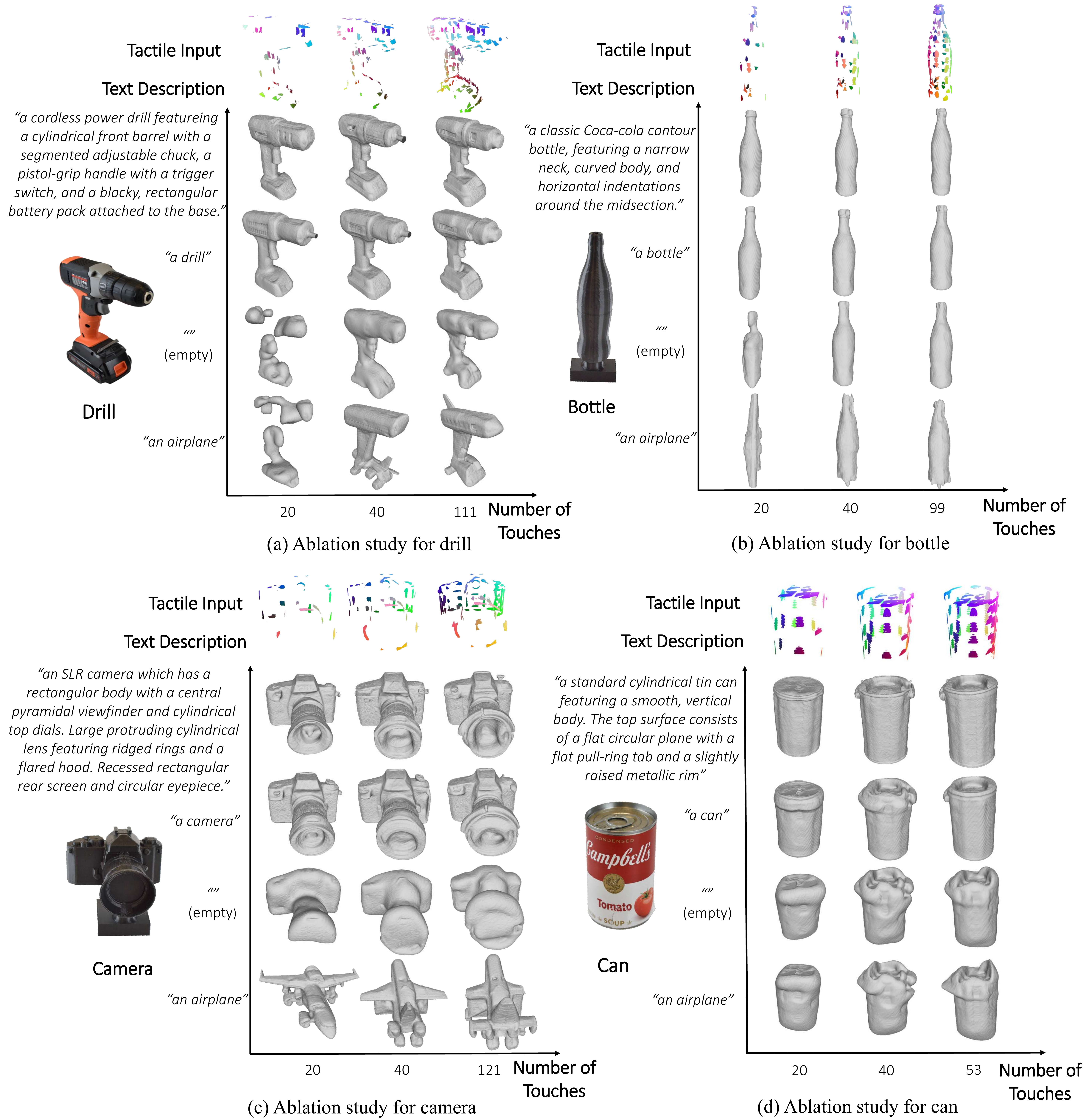}
\caption{We study the effect of the number of tactile measurements and the text description used to guide the diffusion prior on reconstruction quality. Columns correspond to increasing numbers of touches, while rows correspond to different text descriptions (detailed, class-level, empty, and incorrect). Our method reconstructs plausible geometries with as few as 20 touches while more informative descriptions guide the reconstruction toward more realistic shapes.}
\label{fig:ablation}
\end{figure}

\arxiv{fixed image font for full ablation figure ("a can" in different font)}

\subsection{Ablation Studies}
\subsubsection{Ablation: Tactile-Only Reconstruction.} 

\cref{fig:ablationtactonly} shows 3D reconstructions of real objects obtained using tactile observations only, after removing the diffusion prior. The reconstruction quality degrades significantly compared to our full method, highlighting the importance of the diffusion model. This behavior is expected, as tactile sensing provides only local geometric measurements. With a limited number of touches, the reconstruction problem remains highly underconstrained without the global guidance provided by the diffusion prior. 

\subsubsection{Ablation: Text Description and Touch Count.}

We study how the number of touches and the quality of text description affect the reconstruction performance of TouchAnything. To evaluate the effect of additional tactile supervision, we run TouchAnything using $20$, $40$, and all available touches. We also investigate the affect of semantic guidance by varying the text description provided to the model. Specifically, we consider four levels of semantic text guidance: an incorrect description (e.g., “an airplane”), no description (an empty string), a class-level description (e.g., “a camera”), and a highly detailed instance-specific description. \cref{fig:ablation} shows the reconstruction results of TouchAnything under different input combinations on four real-world objects: a camera, drill, cola bottle, and tomato soup can. With only $20$ tactile observations, we show that a class-level text description (e.g., “a camera”) is sufficient for TouchAnything to correctly complete the untouched regions, and a more detailed description does not necessarily improve reconstruction as precise geometric details are hard to specify through text. The influence of the diffusion prior is further illustrated when an incorrect description, such as “an airplane,” is used, which causes the model to hallucinate structures in the unseen regions that match the incorrect semantics. Even when using an empty string as the text description, we show that the diffusion model can still reasonably infer unseen geometry based on common-sense geometric reasoning, such as connecting aligned surface patches or completing partial cylindrical structures. For completeness, we include the quantitative results (EMD) of TouchAnything under different input combinations for four real-world objects with ground-truth meshes in the supplementary material.

\section{Discussion and Conclusion}

We present TouchAnything, a method for reconstructing 3D object geometry from sparse tactile measurements by leveraging an off-the-shelf pretrained 2D diffusion model as a semantic and geometric prior. Our approach combines local geometric constraints derived from tactile sensing with global guidance from a diffusion prior for 3D reconstruction. As a result, it addresses the fundamentally underconstrained nature of reconstructing shapes from sparse physical contacts. Unlike prior tactile reconstruction methods that rely on category-specific training or diffusion models trained directly on tactile datasets, TouchAnything transfers geometric knowledge encoded in large-scale visual diffusion models to the tactile domain. Our results demonstrate that this combination enables accurate and robust 3D reconstruction from only a small number of touches while generalizing to previously unseen objects in an open-world setting. Several promising directions remain for future work. One direction is to investigate the possibility of removing the reliance on text prompts entirely, enabling fully prompt-free reconstruction from tactile measurements. Additionally, our current system assumes that tactile measurements are collected passively. Future work could explore active touch strategies that guide the robot toward the most informative contact locations, enabling more efficient data collection and reconstruction.

\arxiv{typo: "am" to "an"}
\section*{Acknowledgements}

The authors thank Ruohan Zhang and Jingyi Xiang for their help with robot setup and hardware design, thank Ruihan Gao for her discussions on tactile-inspired 3D generation, and thank Yuchen Mo for the support with computing resources. Mohamad Qadri was supported in part by ONR grant N00014-24-1-2272. This work used the Delta system at the National Center for Supercomputing Applications [award OAC 2005572] through allocation CIS240782 from the Advanced Cyberinfrastructure Coordination Ecosystem: Services \& Support (ACCESS) program~\cite{boerner2023access}, which is supported by National Science Foundation grants \#2138259, \#2138286, \#2138307, \#2137603, and \#2138296.

\arxiv{Added Acknowledgements}

\bibliographystyle{splncs04}
\bibliography{main}

@String(CVPR  = {IEEE Conf. Comput. Vis. Pattern Recog.})

@String(NeurIPS = {Adv. Neural Inform. Process. Syst.})

@String(AAAI  = {AAAI})

@String(CVPR  = {CVPR})

@String(NeurIPS = {NeurIPS})

@article{huang2024normalflow,
  title={NormalFlow: Fast, Robust, and Accurate Contact-based Object 6DoF Pose Tracking with Vision-based Tactile Sensors},
  author={Huang, Hung-Jui and Kaess, Michael and Yuan, Wenzhen},
  journal={IEEE Robotics and Automation Letters},
  year={2024},
  publisher={IEEE}
}

@inproceedings{boerner2023access,
  author    = {Boerner, Timothy J. and Deems, Stephen and Furlani, Thomas R. and Knuth, Shelley L. and Towns, John},
  title     = {{ACCESS: Advancing Innovation: NSF’s Advanced Cyberinfrastructure Coordination Ecosystem: Services \& Support}},
  booktitle = {Practice and Experience in Advanced Research Computing ({PEARC} '23)},
  year      = {2023},
  month     = {July},
  address   = {Portland, OR, USA},
  publisher = {{ACM}},
  location  = {New York, NY, USA},
  pages     = {4},
  doi       = {10.1145/3569951.3597559},
  url       = {https://doi.org/10.1145/3569951.3597559}
}

@inproceedings{Ansel_PyTorch_2_Faster_2024,
author = {Ansel, Jason and Yang, Edward and He, Horace and Gimelshein, Natalia and Jain, Animesh and Voznesensky, Michael and Bao, Bin and Bell, Peter and Berard, David and Burovski, Evgeni and Chauhan, Geeta and Chourdia, Anjali and Constable, Will and Desmaison, Alban and DeVito, Zachary and Ellison, Elias and Feng, Will and Gong, Jiong and Gschwind, Michael and Hirsh, Brian and Huang, Sherlock and Kalambarkar, Kshiteej and Kirsch, Laurent and Lazos, Michael and Lezcano, Mario and Liang, Yanbo and Liang, Jason and Lu, Yinghai and Luk, CK and Maher, Bert and Pan, Yunjie and Puhrsch, Christian and Reso, Matthias and Saroufim, Mark and Siraichi, Marcos Yukio and Suk, Helen and Suo, Michael and Tillet, Phil and Wang, Eikan and Wang, Xiaodong and Wen, William and Zhang, Shunting and Zhao, Xu and Zhou, Keren and Zou, Richard and Mathews, Ajit and Chanan, Gregory and Wu, Peng and Chintala, Soumith},
booktitle = {29th ACM International Conference on Architectural Support for Programming Languages and Operating Systems, Volume 2 (ASPLOS '24)},
doi = {10.1145/3620665.3640366},
month = apr,
publisher = {ACM},
title = {{PyTorch 2: Faster Machine Learning Through Dynamic Python Bytecode Transformation and Graph Compilation}},
url = {https://docs.pytorch.org/assets/pytorch2-2.pdf},
year = {2024}
}

@article{Laine2020diffrast,
  title   = {Modular Primitives for High-Performance Differentiable Rendering},
  author  = {Samuli Laine and Janne Hellsten and Tero Karras and Yeongho Seol and Jaakko Lehtinen and Timo Aila},
  journal = {ACM Transactions on Graphics},
  year    = {2020},
  volume  = {39},
  number  = {6}
}

@inproceedings{qadri2022incopt,
  title={InCOpt: Incremental constrained optimization using the Bayes tree},
  author={Qadri, Mohamad and Sodhi, Paloma and Mangelson, Joshua G and Dellaert, Frank and Kaess, Michael},
  booktitle={2022 IEEE/RSJ International Conference on Intelligent Robots and Systems (IROS)},
  pages={6381--6388},
  year={2022},
  organization={IEEE}
}

@inproceedings{qadri2024learning,
  title={Learning covariances for estimation with constrained bilevel optimization},
  author={Qadri, Mohamad and Manchester, Zachary and Kaess, Michael},
  booktitle={2024 IEEE International Conference on Robotics and Automation (ICRA)},
  pages={15951--15957},
  year={2024},
  organization={IEEE}
}

@article{poole2022dreamfusion,
  title={Dreamfusion: Text-to-3d using 2d diffusion},
  author={Poole, Ben and Jain, Ajay and Barron, Jonathan T and Mildenhall, Ben},
  journal={arXiv preprint arXiv:2209.14988},
  year={2022}
}

@article{gao2024tactile,
  title={Tactile DreamFusion: Exploiting tactile sensing for 3D generation},
  author={Gao, Ruihan and Deng, Kangle and Yang, Gengshan and Yuan, Wenzhen and Zhu, Jun-Yan},
  journal={Advances in Neural Information Processing Systems},
  volume={37},
  pages={29839--29863},
  year={2024}
}

@inproceedings{lin2023magic3d,
  title={Magic3d: High-resolution text-to-3d content creation},
  author={Lin, Chen-Hsuan and Gao, Jun and Tang, Luming and Takikawa, Towaki and Zeng, Xiaohui and Huang, Xun and Kreis, Karsten and Fidler, Sanja and Liu, Ming-Yu and Lin, Tsung-Yi},
  booktitle={Proceedings of the IEEE/CVF conference on computer vision and pattern recognition},
  pages={300--309},
  year={2023}
}

@inproceedings{chen2023fantasia3d,
  title={Fantasia3d: Disentangling geometry and appearance for high-quality text-to-3d content creation},
  author={Chen, Rui and Chen, Yongwei and Jiao, Ningxin and Jia, Kui},
  booktitle={Proceedings of the IEEE/CVF international conference on computer vision},
  pages={22246--22256},
  year={2023}
}

@article{wang2023prolificdreamer,
  title={Prolificdreamer: High-fidelity and diverse text-to-3d generation with variational score distillation},
  author={Wang, Zhengyi and Lu, Cheng and Wang, Yikai and Bao, Fan and Li, Chongxuan and Su, Hang and Zhu, Jun},
  journal={Advances in neural information processing systems},
  volume={36},
  pages={8406--8441},
  year={2023}
}

@inproceedings{qiu2024richdreamer,
  title={Richdreamer: A generalizable normal-depth diffusion model for detail richness in text-to-3d},
  author={Qiu, Lingteng and Chen, Guanying and Gu, Xiaodong and Zuo, Qi and Xu, Mutian and Wu, Yushuang and Yuan, Weihao and Dong, Zilong and Bo, Liefeng and Han, Xiaoguang},
  booktitle={Proceedings of the IEEE/CVF conference on computer vision and pattern recognition},
  pages={9914--9925},
  year={2024}
}

@article{kasten2023point,
  title={Point cloud completion with pretrained text-to-image diffusion models},
  author={Kasten, Yoni and Rahamim, Ohad and Chechik, Gal},
  journal={Advances in Neural Information Processing Systems},
  volume={36},
  pages={12171--12191},
  year={2023}
}

@InProceedings{Azinovic_2022_CVPR,
    author    = {Azinovi\'c, Dejan and Martin-Brualla, Ricardo and Goldman, Dan B and Nie{\ss}ner, Matthias and Thies, Justus},
    title     = {Neural RGB-D Surface Reconstruction},
    booktitle = {Proceedings of the IEEE/CVF Conference on Computer Vision and Pattern Recognition (CVPR)},
    month     = {June},
    year      = {2022},
    pages     = {6290-6301}
}

@article{schaefer2024sc,
  title={Sc-diff: 3d shape completion with latent diffusion models},
  author={Schaefer, Simon and Galvis, Juan D and Zuo, Xingxing and Leutengger, Stefan},
  journal={arXiv preprint arXiv:2403.12470},
  year={2024}
}

@inproceedings{shen2021dmtet,
title = {Deep Marching Tetrahedra: a Hybrid Representation for High-Resolution 3D Shape Synthesis},
author = {Tianchang Shen and Jun Gao and Kangxue Yin and Ming-Yu Liu and Sanja Fidler},
year = {2021},
booktitle = {Advances in Neural Information Processing Systems (NeurIPS)}
}

@article{mildenhall2021nerf,
  title={Nerf: Representing scenes as neural radiance fields for view synthesis},
  author={Mildenhall, Ben and Srinivasan, Pratul P and Tancik, Matthew and Barron, Jonathan T and Ramamoorthi, Ravi and Ng, Ren},
  journal={Communications of the ACM},
  volume={65},
  number={1},
  pages={99--106},
  year={2021},
  publisher={ACM New York, NY, USA}
}

@article{shen2021deep,
  title={Deep marching tetrahedra: a hybrid representation for high-resolution 3d shape synthesis},
  author={Shen, Tianchang and Gao, Jun and Yin, Kangxue and Liu, Ming-Yu and Fidler, Sanja},
  journal={Advances in Neural Information Processing Systems},
  volume={34},
  pages={6087--6101},
  year={2021}
}

@article{kerbl20233d,
  title={3d gaussian splatting for real-time radiance field rendering.},
  author={Kerbl, Bernhard and Kopanas, Georgios and Leimk{\"u}hler, Thomas and Drettakis, George and others},
  journal={ACM Trans. Graph.},
  volume={42},
  number={4},
  pages={139--1},
  year={2023}
}

@inproceedings{park2019deepsdf,
  title={Deepsdf: Learning continuous signed distance functions for shape representation},
  author={Park, Jeong Joon and Florence, Peter and Straub, Julian and Newcombe, Richard and Lovegrove, Steven},
  booktitle={Proceedings of the IEEE/CVF conference on computer vision and pattern recognition},
  pages={165--174},
  year={2019}
}

@inproceedings{qadri2023neural,
  title={Neural Implicit Surface Reconstruction using Imaging Sonar},
  author={Qadri, Mohamad and Kaess, Michael and Gkioulekas, Ioannis},
  booktitle={2023 IEEE International Conference on Robotics and Automation (ICRA)},
  pages={1040--1047},
  year={2023},
  organization={IEEE}
}

@inproceedings{qadri2024aoneus,
  title={Aoneus: A neural rendering framework for acoustic-optical sensor fusion},
  author={Qadri, Mohamad and Zhang, Kevin and Hinduja, Akshay and Kaess, Michael and Pediredla, Adithya and Metzler, Christopher A},
  booktitle={ACM SIGGRAPH 2024 Conference Papers},
  pages={1--12},
  year={2024}
}

@article{qu2024z,
  title={Z-splat: Z-axis Gaussian splatting for camera-sonar fusion},
  author={Qu, Ziyuan and Vengurlekar, Omkar and Qadri, Mohamad and Zhang, Kevin and Kaess, Michael and Metzler, Christopher and Jayasuriya, Suren and Pediredla, Adithya},
  journal={IEEE transactions on pattern analysis and machine intelligence},
  volume={47},
  number={9},
  pages={7255--7267},
  year={2024},
  publisher={IEEE}
}

@inproceedings{lin2025acoustic,
  title={Acoustic neural 3d reconstruction under pose drift},
  author={Lin, Tianxiang and Qadri, Mohamad and Zhang, Kevin and Pediredla, Adithya and Metzler, Christopher A and Kaess, Michael},
  booktitle={2025 IEEE/RSJ International Conference on Intelligent Robots and Systems (IROS)},
  pages={12704--12711},
  year={2025},
  organization={IEEE}
}

@inproceedings{rafidashti2025neuradar,
  title={Neuradar: Neural radiance fields for automotive radar point clouds},
  author={Rafidashti, Mahan and Lan, Ji and Fatemi, Maryam and Fu, Junsheng and Hammarstrand, Lars and Svensson, Lennart},
  booktitle={Proceedings of the Computer Vision and Pattern Recognition Conference},
  pages={2488--2498},
  year={2025}
}

@inproceedings{kung2025radarsplat,
  title={Radarsplat: Radar gaussian splatting for high-fidelity data synthesis and 3d reconstruction of autonomous driving scenes},
  author={Kung, Pou-Chun and Harisha, Skanda and Vasudevan, Ram and Eid, Aline and Skinner, Katherine A},
  booktitle={Proceedings of the IEEE/CVF International Conference on Computer Vision},
  pages={27596--27606},
  year={2025}
}

@inproceedings{zhao2024tclc,
  title={Tclc-gs: Tightly coupled lidar-camera gaussian splatting for autonomous driving: Supplementary materials},
  author={Zhao, Cheng and Sun, Su and Wang, Ruoyu and Guo, Yuliang and Wan, Jun-Jun and Huang, Zhou and Huang, Xinyu and Chen, Yingjie Victor and Ren, Liu},
  booktitle={European Conference on Computer Vision},
  pages={91--106},
  year={2024},
  organization={Springer}
}

@article{comi2024touchsdf,
  title={Touchsdf: A deepsdf approach for 3d shape reconstruction using vision-based tactile sensing},
  author={Comi, Mauro and Lin, Yijiong and Church, Alex and Tonioni, Alessio and Aitchison, Laurence and Lepora, Nathan F},
  journal={IEEE Robotics and Automation Letters},
  volume={9},
  number={6},
  pages={5719--5726},
  year={2024},
  publisher={IEEE}
}

@article{suresh2024neuralfeels,
  title={NeuralFeels with neural fields: Visuotactile perception for in-hand manipulation},
  author={Suresh, Sudharshan and Qi, Haozhi and Wu, Tingfan and Fan, Taosha and Pineda, Luis and Lambeta, Mike and Malik, Jitendra and Kalakrishnan, Mrinal and Calandra, Roberto and Kaess, Michael and others},
  journal={Science Robotics},
  volume={9},
  number={96},
  pages={eadl0628},
  year={2024},
  publisher={American Association for the Advancement of Science}
}

@inproceedings{niemeyer2022regnerf,
  title={Regnerf: Regularizing neural radiance fields for view synthesis from sparse inputs},
  author={Niemeyer, Michael and Barron, Jonathan T and Mildenhall, Ben and Sajjadi, Mehdi SM and Geiger, Andreas and Radwan, Noha},
  booktitle={Proceedings of the IEEE/CVF conference on computer vision and pattern recognition},
  pages={5480--5490},
  year={2022}
}

@inproceedings{wang2023sparsenerf,
  title={Sparsenerf: Distilling depth ranking for few-shot novel view synthesis},
  author={Wang, Guangcong and Chen, Zhaoxi and Loy, Chen Change and Liu, Ziwei},
  booktitle={Proceedings of the IEEE/CVF international conference on computer vision},
  pages={9065--9076},
  year={2023}
}

@techreport{shapenet2015,
  title       = {{ShapeNet: An Information-Rich 3D Model Repository}},
  author      = {Chang, Angel X. and Funkhouser, Thomas and Guibas, Leonidas and Hanrahan, Pat and Huang, Qixing and Li, Zimo and Savarese, Silvio and Savva, Manolis and Song, Shuran and Su, Hao and Xiao, Jianxiong and Yi, Li and Yu, Fisher},
  number      = {arXiv:1512.03012 [cs.GR]},
  institution = {Stanford University --- Princeton University --- Toyota Technological Institute at Chicago},
  year        = {2015}
}

@inproceedings{chibane2020implicit,
  title={Implicit feature networks for texture completion from partial 3d data},
  author={Chibane, Julian and Pons-Moll, Gerard},
  booktitle={European Conference on Computer Vision},
  pages={717--725},
  year={2020},
  organization={Springer}
}

@inproceedings{yuan2018pcn,
  title={Pcn: Point completion network},
  author={Yuan, Wentao and Khot, Tejas and Held, David and Mertz, Christoph and Hebert, Martial},
  booktitle={2018 international conference on 3D vision (3DV)},
  pages={728--737},
  year={2018},
  organization={IEEE}
}

@article{instantngp,
author = {M\"{u}ller, Thomas and Evans, Alex and Schied, Christoph and Keller, Alexander},
title = {Instant neural graphics primitives with a multiresolution hash encoding},
year = {2022},
issue_date = {July 2022},
publisher = {Association for Computing Machinery},
address = {New York, NY, USA},
volume = {41},
number = {4},
issn = {0730-0301},
url = {https://doi.org/10.1145/3528223.3530127},
doi = {10.1145/3528223.3530127},
journal = {ACM Trans. Graph.},
month = jul,
articleno = {102},
numpages = {15}
}

@article{huang2025gelslam,
  title={Gelslam: A real-time, high-fidelity, and robust 3D tactile slam system},
  author={Huang, Hung-Jui and Mirzaee, Mohammad Amin and Kaess, Michael and Yuan, Wenzhen},
  journal={arXiv preprint arXiv:2508.15990},
  year={2025}
}

@inproceedings{wang2025touch2shape,
  title={Touch2Shape: Touch-Conditioned 3D Diffusion for Shape Exploration and Reconstruction},
  author={Wang, Yuanbo and Zhang, Zhaoxuan and Qiu, Jiajin and Sun, Dilong and Meng, Zhengyu and Wei, Xiaopeng and Yang, Xin},
  booktitle={Proceedings of the Computer Vision and Pattern Recognition Conference},
  pages={5656--5665},
  year={2025}
}

@article{zhang2025end,
  title={End-to-End Diffusion-Based 3D Object Reconstruction From Robotic Tactile Sensing},
  author={Zhang, Han and Zhang, Xiaohui and Huang, Jun and Feng, Zhao and Xiao, Xiaohui},
  journal={IEEE Robotics and Automation Letters},
  volume={11},
  number={2},
  pages={1434--1441},
  year={2025},
  publisher={IEEE}
}

@inproceedings{comi2025snap,
  title={Snap-it, tap-it, splat-it: Tactile-informed 3d gaussian splatting for reconstructing challenging surfaces},
  author={Comi, Mauro and Tonioni, Alessio and Tremblay, Jonathan and Yang, Max and Blukis, Valts and Lin, Yijiong and Lepora, Nathan F and Aitchison, Laurence},
  booktitle={2025 International Conference on 3D Vision (3DV)},
  pages={1134--1143},
  year={2025},
  organization={IEEE}
}

@inproceedings{swann2024touch,
  title={Touch-gs: Visual-tactile supervised 3d gaussian splatting},
  author={Swann, Aiden and Strong, Matthew and Do, Won Kyung and Camps, Gadiel Sznaier and Schwager, Mac and Kennedy, Monroe},
  booktitle={2024 IEEE/RSJ International Conference on Intelligent Robots and Systems (IROS)},
  pages={10511--10518},
  year={2024},
  organization={IEEE}
}

@inproceedings{fang2025fusionsense,
  title={Fusionsense: Bridging common sense, vision, and touch for robust sparse-view reconstruction},
  author={Fang, Irving and Shi, Kairui and He, Xujin and Tan, Siqi and Wang, Yifan and Zhao, Hanwen and Huang, Hung-Jui and Yuan, Wenzhen and Feng, Chen and Zhang, Jing},
  booktitle={2025 IEEE International Conference on Robotics and Automation (ICRA)},
  pages={15798--15805},
  year={2025},
  organization={IEEE}
}

@inproceedings{suresh2022shapemap,
  title={Shapemap 3-d: Efficient shape mapping through dense touch and vision},
  author={Suresh, Sudharshan and Si, Zilin and Mangelson, Joshua G and Yuan, Wenzhen and Kaess, Michael},
  booktitle={2022 International Conference on Robotics and Automation (ICRA)},
  pages={7073--7080},
  year={2022},
  organization={IEEE}
}

@ARTICLE{si2022taxim,
  author={Si, Zilin and Yuan, Wenzhen},
  journal={IEEE Robotics and Automation Letters}, 
  title={Taxim: An Example-Based Simulation Model for GelSight Tactile Sensors}, 
  year={2022},
  volume={7},
  number={2},
  pages={2361-2368},
  doi={10.1109/LRA.2022.3142412}}

@article{yuksel2015poissondisksampling,
author = {Yuksel, Cem},
title = {Sample Elimination for Generating Poisson Disk Sample Sets},
year = {2015},
issue_date = {May 2015},
publisher = {The Eurographs Association \& John Wiley \& Sons, Ltd.},
address = {Chichester, GBR},
volume = {34},
number = {2},
issn = {0167-7055},
url = {https://doi.org/10.1111/cgf.12538},
doi = {10.1111/cgf.12538},
journal = {Comput. Graph. Forum},
month = may,
pages = {25–32},
numpages = {8}
}

@article{zhou2018open3d,
   author  = {Qian-Yi Zhou and Jaesik Park and Vladlen Koltun},
   title   = {{Open3D}: {A} Modern Library for {3D} Data Processing},
   journal = {arXiv:1801.09847},
   year    = {2018},
}

@InProceedings{ronneberger2015unet,
author="Ronneberger, Olaf
and Fischer, Philipp
and Brox, Thomas",
editor="Navab, Nassir
and Hornegger, Joachim
and Wells, William M.
and Frangi, Alejandro F.",
title="U-Net: Convolutional Networks for Biomedical Image Segmentation",
booktitle="Medical Image Computing and Computer-Assisted Intervention -- MICCAI 2015",
year="2015",
publisher="Springer International Publishing",
address="Cham",
pages="234--241",
isbn="978-3-319-24574-4"
}

@article{yuan2017gelsight,
  title={Gelsight: High-resolution robot tactile sensors for estimating geometry and force},
  author={Yuan, Wenzhen and Dong, Siyuan and Adelson, Edward H},
  journal={Sensors},
  volume={17},
  number={12},
  pages={2762},
  year={2017},
  publisher={MDPI}
}

@INPROCEEDINGS{wang2021wedge,
  author={Wang, Shaoxiong and She, Yu and Romero, Branden and Adelson, Edward},
  booktitle={2021 IEEE International Conference on Robotics and Automation (ICRA)}, 
  title={GelSight Wedge: Measuring High-Resolution 3D Contact Geometry with a Compact Robot Finger}, 
  year={2021},
  volume={},
  number={},
  pages={6468-6475},
  doi={10.1109/ICRA48506.2021.9560783}}

@InProceedings{Rombach_2022_CVPR,
    author    = {Rombach, Robin and Blattmann, Andreas and Lorenz, Dominik and Esser, Patrick and Ommer, Bj\"orn},
    title     = {High-Resolution Image Synthesis With Latent Diffusion Models},
    booktitle = {Proceedings of the IEEE/CVF Conference on Computer Vision and Pattern Recognition (CVPR)},
    month     = {June},
    year      = {2022},
    pages     = {10684-10695}
}

@article{ycb2017,
author = {Berk Calli and Arjun Singh and James Bruce and Aaron Walsman and Kurt Konolige and Siddhartha Srinivasa and Pieter Abbeel and Aaron M Dollar},
title ={Yale-CMU-Berkeley dataset for robotic manipulation research},
journal = {The International Journal of Robotics Research},
volume = {36},
number = {3},
pages = {261-268},
year = {2017},
doi = {10.1177/0278364917700714},
URL = { 
        https://doi.org/10.1177/0278364917700714
},
eprint = { 
        https://doi.org/10.1177/0278364917700714
}
}

@article{ravi2020pytorch3d,
    author = {Nikhila Ravi and Jeremy Reizenstein and David Novotny and Taylor Gordon
                  and Wan-Yen Lo and Justin Johnson and Georgia Gkioxari},
    title = {Accelerating 3D Deep Learning with PyTorch3D},
    journal = {arXiv:2007.08501},
    year = {2020},
}

@inproceedings{li2023neuralangelo,
  title={Neuralangelo: High-Fidelity Neural Surface Reconstruction},
  author={Li, Zhaoshuo and M\"uller, Thomas and Evans, Alex and Taylor, Russell H and Unberath, Mathias and Liu, Ming-Yu and Lin, Chen-Hsuan},
  booktitle={IEEE Conference on Computer Vision and Pattern Recognition ({CVPR})},
  year={2023}
}

@inproceedings{wu2024reconfusion,
  title={Reconfusion: 3d reconstruction with diffusion priors},
  author={Wu, Rundi and Mildenhall, Ben and Henzler, Philipp and Park, Keunhong and Gao, Ruiqi and Watson, Daniel and Srinivasan, Pratul P and Verbin, Dor and Barron, Jonathan T and Poole, Ben and others},
  booktitle={Proceedings of the IEEE/CVF conference on computer vision and pattern recognition},
  pages={21551--21561},
  year={2024}
}

@inproceedings{zou2024sparse3d,
  title={Sparse3d: Distilling multiview-consistent diffusion for object reconstruction from sparse views},
  author={Zou, Zixin and Cheng, Weihao and Cao, Yan-Pei and Huang, Shi-Sheng and Shan, Ying and Zhang, Song-Hai},
  booktitle={Proceedings of the AAAI conference on artificial intelligence},
  volume={38},
  number={7},
  pages={7900--7908},
  year={2024}
}

@InProceedings{Ni_2025_CVPR,
    author    = {Ni, Junfeng and Liu, Yu and Lu, Ruijie and Zhou, Zirui and Zhu, Song-Chun and Chen, Yixin and Huang, Siyuan},
    title     = {Decompositional Neural Scene Reconstruction with Generative Diffusion Prior},
    booktitle = {Proceedings of the IEEE/CVF Conference on Computer Vision and Pattern Recognition (CVPR)},
    month     = {June},
    year      = {2025},
    pages     = {6022-6033}
}

\clearpage
\appendix

\begin{center}
    {\Large \bfseries Supplementary Material \par}
\end{center}

In this supplementary material, we provide additional details regarding our methodology, experiments and results.

\section{Data Collection}

\subsection{Simulation Data Collection}
\subsubsection{Tactile image generation}

\begin{figure}[!h]
\centering
\includegraphics[width=\linewidth]{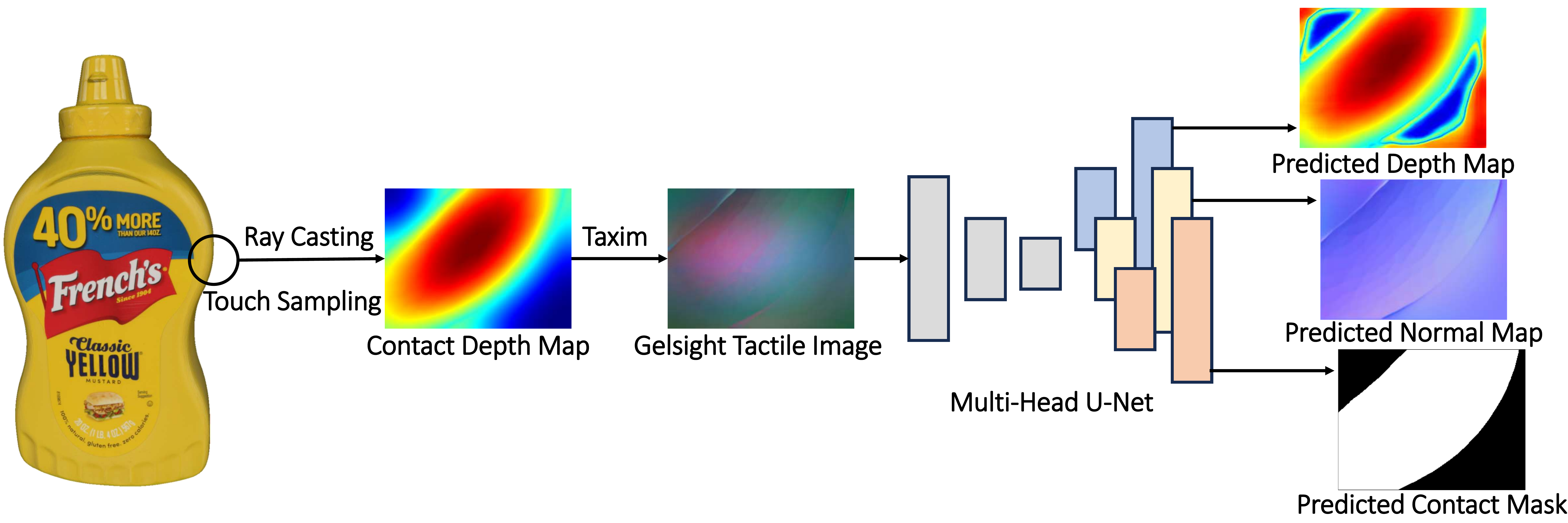}
\caption{A Multi-Head U-Net is implemented to derive local geometry from Gelsight Tactile Images.}
\label{fig:simulation_geometry}
\end{figure}


We propose a physically grounded simulation pipeline for tactile data generation that captures realistic contact mechanics by leveraging Taxim~\cite{si2022taxim}, an example-based Gelsight simulation model. The data acquisition process begins with the selection of potential contact points via Poisson Disk Sampling~\cite{yuksel2015poissondisksampling} over the object mesh; here, the surface is intentionally oversampled to ensure high-quality tactile image generation by providing a buffer against unsuitable geometries, such as overly flat regions or extreme depressions. 

Once these candidates are established, the sensor poses are initialized by translating a set distance from each pre-selected contact point along the radial vector originating from the coordinate center. From these initial positions, the system perceives the local surface normals within the sensor's field of view. These sensors are subsequently aligned to the identified normals and driven into the surface at a predefined pressing depth to simulate physical contact. To synthesize the final output, a contact depth map is generated using Open3D ray casting~\cite{zhou2018open3d}, which is then transformed into a $320 \times 240$ pixel Gelsight tactile image through the Taxim framework. Finally, the samples with insufficient contact area are discarded. The sensor poses for the validated tactile contacts are stored.

For our simulation experiments conducted with ShapeNetCore.V2~\cite{shapenet2015} objects, we rescale them to 20\% of their original size following TouchSDF~\cite{comi2024touchsdf}. This makes their sizes comparable with real-world sensor and objects.


\subsubsection{Tactile image to local geometry} To reconstruct the local surface geometry from raw sensory data, we implemented a Multi-Head U-Net architecture~\cite{ronneberger2015unet}. This model serves as a perception module that maps a single RGB GelSight tactile image to three distinct geometric representations: a depth map, a normal map, and a contact mask, as illustrated in Figure~\ref{fig:simulation_geometry}.

The network features a shared encoder to extract common tactile features, followed by three independent decoder heads tailored for each modality. We trained the model on a large-scale synthetic dataset comprising 20k tactile samples. These samples were generated using the Taxim simulation framework based on 78 YCB objects with varying mesh resolutions (16k and 64k). Each tactile sample set includes a ground truth depth map, a ground truth normal map, and a ground truth contact mask.

The model is optimized end-to-end using a weighted multi-task loss function $\mathcal{L}_{total}$, which balances the convergence across different tasks:
\begin{equation}
    \label{unet_loss}
    \mathcal{L}_{total} = \lambda_m \mathcal{L}_{mask} + \lambda_d \mathcal{L}_{depth} + \lambda_n \mathcal{L}_{normal}
\end{equation}
where we employ Binary Cross-Entropy (BCE) for the contact mask loss $\mathcal{L}_{mask}$, $L_1$ loss for the depth loss $\mathcal{L}_{depth}$, and Cosine Similarity for the normal loss $\mathcal{L}_{normal}$.

For TouchAnything, we specifically leverage the predicted depth map and contact mask to synthesize virtual camera observations. The predicted normal maps, while currently redundant for our immediate geometry-derivation stage, are generated by our multi-head architecture to provide a more comprehensive geometric prediction and to facilitate alternative tactile sensing tasks in future research.

\begin{figure}[!h]
\centering
\includegraphics[width=0.7\linewidth]{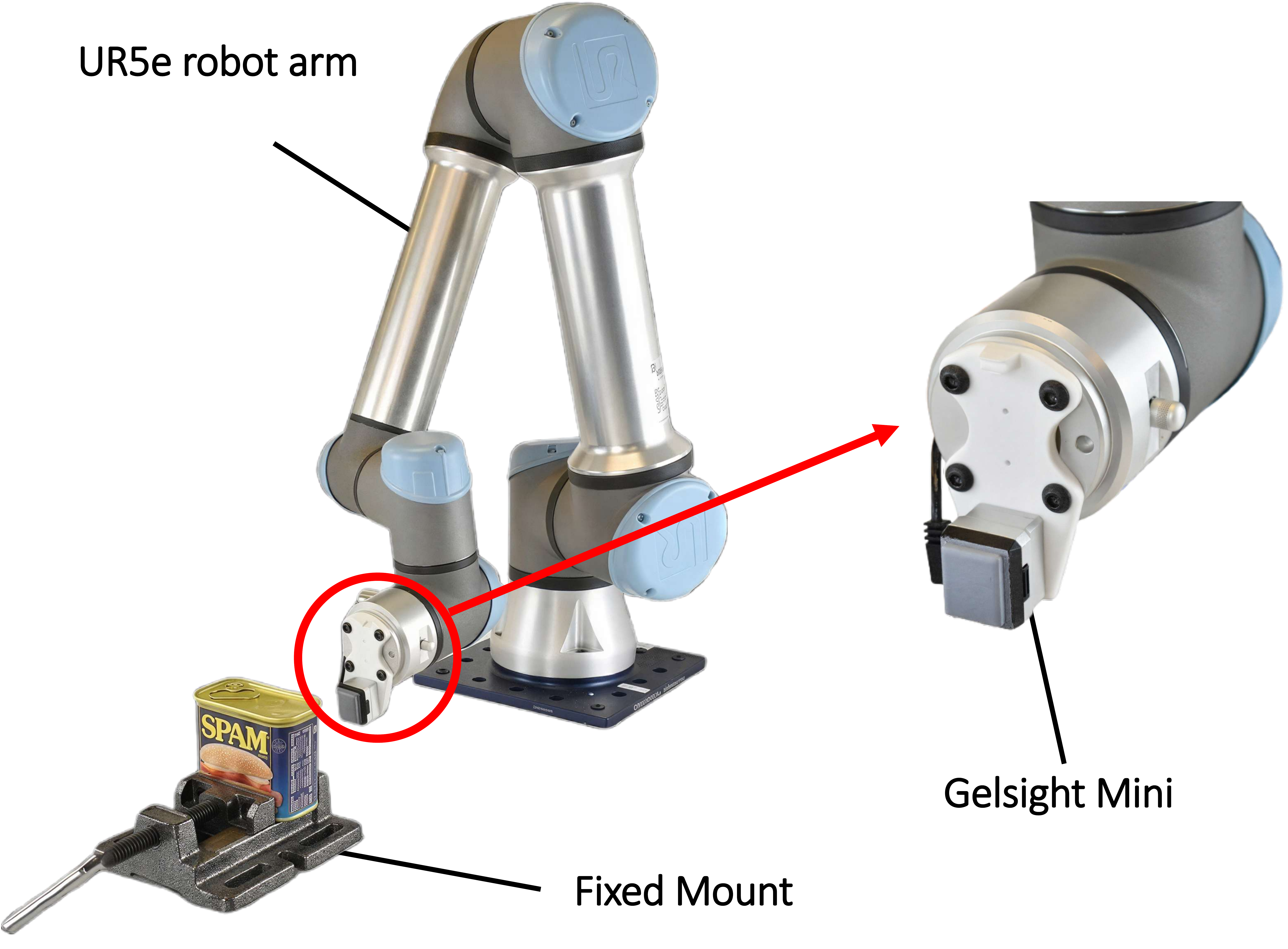}
\caption{Our real-world experiment hardware setup, including a UR5e robot arm, a Gelsight Mini tactile sensor and a fixed mount.}
\label{fig:hardware_setup}
\end{figure}

\subsection{Real-world Data Collection}
Our real-world data collection system includes a 6-DOF UR5e robot arm, a Gelsight Mini tactile sensor and a fixed mount on the table, as shown in Figure~\ref{fig:hardware_setup}. To collect tactile data, we operate the robot arm manually and press the tactile sensor on the surface of the object. A GelSight image is recorded after every contact, and the sensor pose is calculated through the forward kinematics of the robotic arm.

\section{Extended Analysis of Fine Geometry Learning}
\begin{figure}[!ht]
    \centering
    \includegraphics[width=\linewidth]{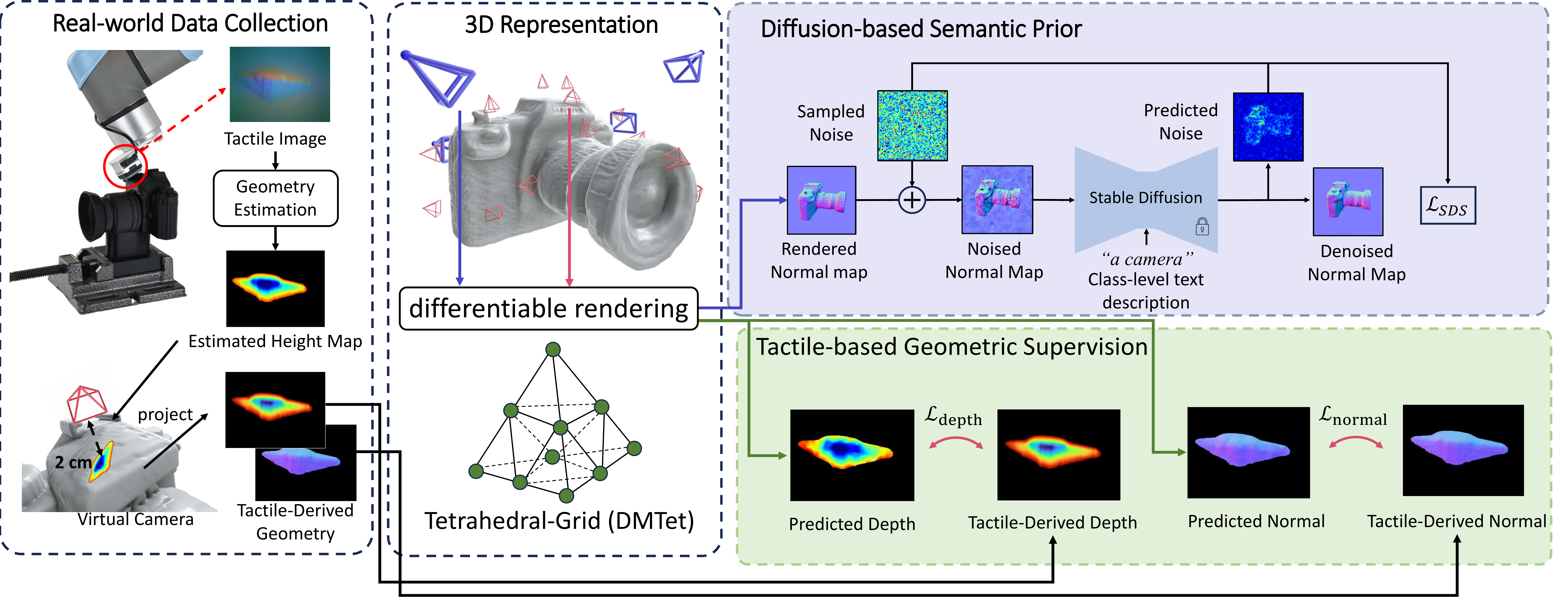}
    \caption{Overview of the stage 2 fine geometry learning pipeline.}
    \label{fig:supp_pipeline}
\end{figure}

In this section, we provide additional visual evidence and implementation details for the geometry refinement stage. The pipeline for the explicit DMTet~\cite{shen2021dmtet}-based refinement is illustrated in Figure~\ref{fig:supp_pipeline}. Utilizing DMTet representation and differentiable mesh renderer~\cite{Laine2020diffrast}, we achieve extremely fast rendering at a very high resolution. This architectural shift is the key for the diffusion model to capture high-frequency geometric details and provide rich texture to the surface.

\begin{figure}[!h]
    \centering
    \includegraphics[width=\linewidth]{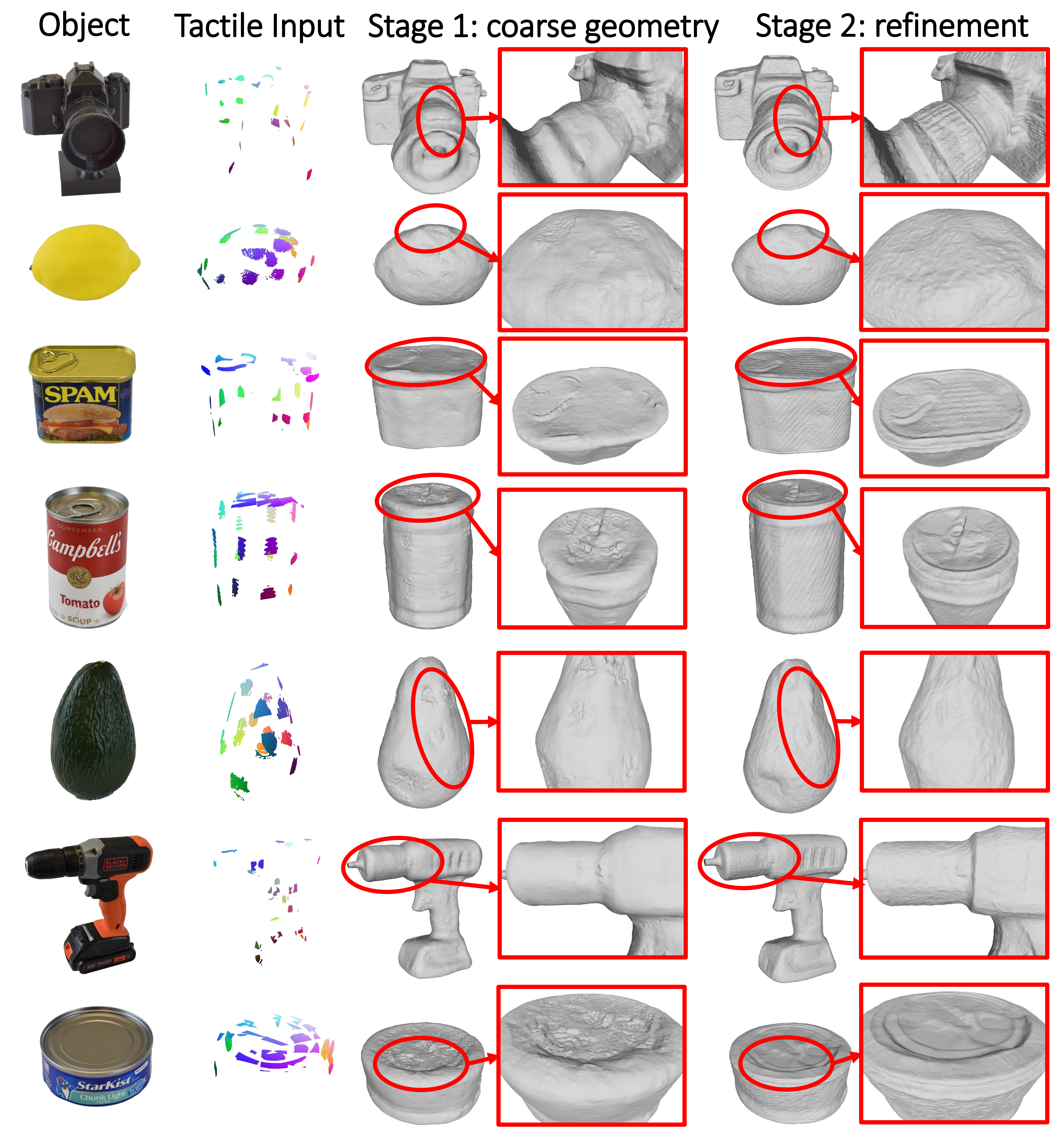}
    \caption{Qualitative comparison between the results before and after fine geometry learning. We choose some representative objects to show how their surface texture evolves after the stage 2 geometry refinement.}
    \label{fig:refinement_comparison}
\end{figure}

To demonstrate the effectiveness of our refinement strategy, we present a qualitative comparison in Figure~\ref{fig:refinement_comparison}.  We choose some representative objects with rich surface details themselves to show how their surface texture evolves after the stage 2 geometry refinement step. While the coarse stage successfully recovers the global topology and basic structure, it often suffers from plain surface texture and lacks fine-grained details because the rendered normal images passed to the diffusion prior are of low resolution. In contrast, after our refinement, the reconstructed results all exhibit fine surface details, such as the rugged texture on the surface of the avocado, the concentric ridges on the top of the can, and the parallel fluting on the lens of the camera. These results provide compelling evidence that our refinement stage effectively recovers high-frequency surface details, leading to more fine-grained object reconstructions.

\section{Quantitative Evaluation for Real-world Reconstruction Results}

\begin{figure}[!h]
    \centering
    \includegraphics[width=\linewidth]{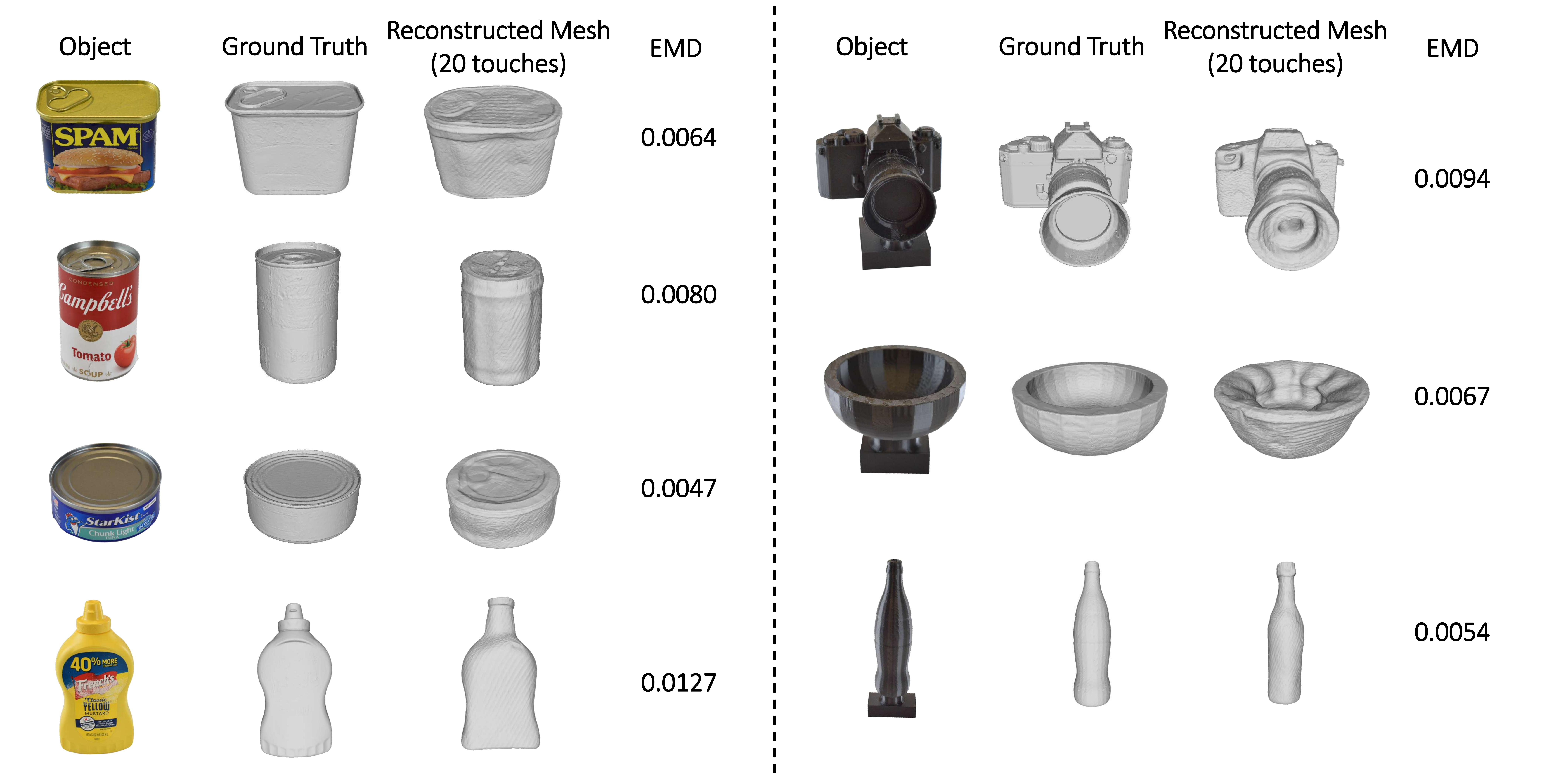}
    \caption{Quantitative evaluation for real-world results.}
    \label{fig:real_world_metrics}
\end{figure}

We present the quantitative metrics for the real-world reconstruction results of selected objects with ground truth meshes. These include the potted meat can, the tomato soup can, the tuna can and the mustard bottle from the YCB dataset~\cite{ycb2017} which have ground truth meshes provided by high-resolution scanner, and the printed camera model, the printed bottle model and the printed bowl model, whose ground truth meshes are from the ShapeNet-Core.V2~\cite{shapenet2015} dataset. They are evaluated using Earth Mover's Distance (EMD). The result is reported in Figure~\ref{fig:real_world_metrics}.

For our ablation study, to further analyze the impact of tactile sample density and semantic guidance, we provide comprehensive Earth Mover's Distance (EMD) metrics for four representative real-world objects: a 3D-printed bottle, a 3D-printed camera, a tomato soup can, and a potted meat can. As shown in Table~\ref{tab:emd_ablation_full}, we evaluate the reconstruction error across varying numbers of robot touches and four levels of text prompts ($p_1$: incorrect, $p_2$: empty, $p_3$: class-level, and $p_4$: detailed). The details of text prompts used for each real-world object is listed in Table~\ref{tab:prompt_details}.

\begin{table}[!h]
    \centering
    \caption{Detailed EMD results for the ablation study, including four representative real-world objects: a 3D-printed bottle, a 3D-printed camera, a tomato soup can, and a potted meat can.}
    \label{tab:emd_ablation_full}

    \begin{subtable}{0.48\textwidth}
        \centering
        \caption{3D-Printed Bottle}
        \begin{tabular}{@{}lccc@{}}
            \toprule
            \diagbox[width=5.5em, height=2.5em]{Prompt}{Touches} & 20 & 40 & 99 \\ \midrule
            $p_1$ & 0.0105 & 0.0067 & 0.0075 \\
            $p_2$ & 0.0058 & 0.0058 & 0.0080 \\
            $p_3$ & 0.0057 & 0.0052 & 0.0066 \\
            $p_4$ & 0.0089 & 0.0050 & 0.0056 \\ \bottomrule
        \end{tabular}
    \end{subtable}
    \hfill
    \begin{subtable}{0.48\textwidth}
        \centering
        \caption{3D-Printed Camera}
        \begin{tabular}{@{}lccc@{}}
            \toprule
            \diagbox[width=5.5em, height=2.5em]{Prompt}{Touches} & 20 & 40 & 121 \\ \midrule
            $p_1$ & 0.0180 & 0.0190 & 0.0221 \\
            $p_2$ & 0.0104 & 0.0071 & 0.0112 \\
            $p_3$ & 0.0098 & 0.0095 & 0.0109 \\
            $p_4$ & 0.0097 & 0.0094 & 0.0100 \\ \bottomrule
        \end{tabular}
    \end{subtable}

    \vspace{1.5em}

    \begin{subtable}{0.48\textwidth}
        \centering
        \caption{Tomato Soup Can}
        \begin{tabular}{@{}lccc@{}}
            \toprule
            \diagbox[width=5.5em, height=2.5em]{Prompt}{Touches} & 20 & 40 & 53 \\ \midrule
            $p_1$ & 0.0099 & 0.0090 & 0.0090 \\
            $p_2$ & 0.0089 & 0.0088 & 0.0100 \\
            $p_3$ & 0.0077 & 0.0082 & 0.0087 \\
            $p_4$ & 0.0079 & 0.0069 & 0.0079 \\ \bottomrule
        \end{tabular}
    \end{subtable}
    \hfill
    \begin{subtable}{0.48\textwidth}
        \centering
        \caption{Potted Meat Can}
        \begin{tabular}{@{}lccc@{}}
            \toprule
            \diagbox[width=5.5em, height=2.5em]{Prompt}{Touches} & 20 & 40 & 91 \\ \midrule
            $p_1$ & 0.0092 & 0.0080 & 0.0082 \\
            $p_2$ & 0.0086 & 0.0086 & 0.0081 \\
            $p_3$ & 0.0062 & 0.0069 & 0.0074 \\
            $p_4$ & 0.0064 & 0.0071 & 0.0080 \\ \bottomrule
        \end{tabular}
    \end{subtable}
\end{table}

The results highlight two key insights regarding the synergy between tactile evidence and semantic priors. First, the quality of the text prompt plays an important role in reconstruction quality. Across most test cases, the class-level ($p_3$) and detailed ($p_4$) descriptions consistently outperform the incorrect ($p_1$) or empty ($p_2$) prompts. This demonstrates that a correct semantic prior is essential for the diffusion model to resolve the inherent ambiguity of sparse tactile points and to complete untouched regions with category-specific geometric information.

Second, we observe that reconstructions using 20 or 40 touches occasionally yield lower EMD values than those using the full tactile dataset. We attribute this to the non-uniform quality of tactile measurements in dense sampling. During full-scale data collection, the sensor inevitably interacts with highly complex or sharp features, such as the thin metal ring edge on a can or the narrow edges of a camera lens hood. Due to the elastic deformation of the tactile sensor, depth estimation for these thin structures often suffers from a low-pass filtering effect, resulting in thickened artifacts that deviate from the ground truth mesh. In contrast, with fewer but cleaner samples used as local constraints, our pipeline leverages the powerful geometric priors guided by accurate prompts ($p_3, p_4$) to reconstruct plausible structures. 

\begin{table}[h]
    \centering
    \caption{Details of text prompts ($p_1$ to $p_4$) used for each real-world object in the ablation study.}
    \label{tab:prompt_details}
    \small
    \begin{tabular}{@{}llp{8cm}@{}}
        \toprule
        Object & ID & Text Prompt Content \\ \midrule
        \textbf{3D-Printed Bottle} & $p_1$ & an airplane \\
         & $p_2$ & [Empty String] \\
         & $p_3$ & a bottle \\
         & $p_4$ & a classic Coca-cola contour bottle, featuring a narrow neck, curved body, and horizontal indentations around the midsection \\ \midrule
        \textbf{3D-Printed Camera} & $p_1$ & an airplane \\
         & $p_2$ & [Empty String] \\
         & $p_3$ & a camera \\
         & $p_4$ & an SLR camera which has a rectangular body with a central pyramidal viewfinder and cylindrical top dials. Large protruding cylindrical lens featuring ridged rings and a flared hood. Recessed rectangular rear screen and circular eyepiece. \\ \midrule
         \textbf{Tomato Soup Can} & $p_1$ & an airplane \\
         & $p_2$ & [Empty String] \\
         & $p_3$ & a can \\
         & $p_4$ & a standard cylindrical tin can featuring a smooth, vertical body. The top surface consists of a flat circular plane with a flat pull-ring tab and a slightly raised metallic rim \\ \midrule
        \textbf{Potted Meat Can} & $p_1$ & an airplane \\
         & $p_2$ & [Empty String] \\
         & $p_3$ & a can \\
         & $p_4$ & a rectangular cuboid can with heavily rounded vertical edges. The top surface features a raised lip and a flat central plane. An integrated loop-shaped pull-tab attached rests flat on the top surface. \\ \bottomrule
        
    \end{tabular}
\end{table}
\clearpage
\section{Implementation Details}

\subsection{3D Representation Implementation Details}
We implement TouchAnything using PyTorch~\cite{Ansel_PyTorch_2_Faster_2024} with 16-bit mixed precision. All models are optimized using the Adam optimizer. We set the learning rate to $1 \times 10^{-2}$ for the hash grid and $1 \times 10^{-3}$ for the SDF network and DMTet parameters. We employ the pretrained Stable Diffusion v2.1-base model~\cite{Rombach_2022_CVPR} as the SDS backbone. The optimization is split into two distinct stages as follows.

\subsubsection{Stage 1: Coarse Geometry Optimization}
For the implicit geometry in Stage 1, we represent the SDF using a multi-resolution hash-grid encoder followed by a shallow MLP, following the implementation of Instant-NGP~\cite{instantngp} and Neuralangelo~\cite{li2023neuralangelo}. The encoder uses a Progressive Hash Grid~\cite{li2023neuralangelo} with 16 levels, 2 features per level, and a maximum hash table size of $2^{20}$. The base resolution is set to 16 with a per-level growth scale of approximately $1.45$. The levels are incrementally unlocked starting from level 8 and updates every 400 steps from step 1000. The SDF network is implemented as Vanilla MLPs with one hidden layer of 64 neurons and ReLU activations.

We adopt the NeuS volume renderer~\cite{qadri2024aoneus} for stage 1. For tactile-derived supervision, we sample with $16384$ rays in a batch, and sample 512 points along each ray to compute the geometric constraints. For diffusion-based guidance, we sample $8$ random camera views in a batch and input the rendered normal maps to the diffusion model.

In Stage 1, we make the normal loss weight $\lambda_{normal}$ gradually increase to its target value (from 0.025 to 1.0 in simulation, and from 0.1 to 4.0 in real-world experiments) following a linear schedule over the first 6,000 steps. This progressive strategy prevents the normal supervision from disrupting the global shape optimization during the initial phase, allowing the model to prioritize coarse geometry establishment via depth constraints before shifting focus toward the refinement of fine-grained surface details in the later stages.

\subsubsection{Stage 2: Fine Geometry Refinement}
Stage 2 utilizes an explicit tetrahedral grid (DMTet) with a resolution of 256, initialized from the optimized Stage 1 SDF. For surface extraction we use marching tetrahedra where the extraction threshold is set to $-0.03$ for simulation objects and $0.0$ for real-world objects.

We use nvdiffrast~\cite{Laine2020diffrast} as the differentiable renderer for stage 2. For tactile-derived supervision, we sample $\min(\text{number of touches, 32})$ sets of tactile-derived images to compute the geometric constraints. For diffusion-based guidance, we sample $4$ random camera views in a batch and input the rendered normal maps to the diffusion model.

\arxiv{removed table 4}

\subsection{Multi-Head U-Net Implementation Details}
The geometry estimation module for simulated tactile data is implemented as a Multi-Head U-Net~\cite{ronneberger2015unet} designed to map a GelSight tactile image $T \in \mathbb{R}^{3 \times H \times W}$ to three distinct geometric representations. The network follows an encoder-decoder paradigm with a shared feature extractor and task-specific branches.

\subsubsection{Shared Encoder}
The encoder consists of a series of blocks designed to extract hierarchical tactile features. It begins with an initial double convolution (\textit{DoubleConv}) layer that maps the input to 64 channels. This is followed by four downsampling stages (\textit{Down}), each utilizing a $2 \times 2$ max-pooling operation followed by a \textit{DoubleConv} block. The channel dimensions increase progressively as $\{64, 128, 256, 512, 1024\}$, effectively capturing both local texture and global contact structure. Each \textit{DoubleConv} block consists of two $3 \times 3$ convolutions, each followed by Batch Normalization and ReLU activation.

\subsubsection{Multi-Task Decoders}
To reconstruct local geometry, the shared latent features are passed into three independent decoder branches:
\begin{itemize}
    \item \textbf{Depth Decoder:} Predicts a single-channel depth map $\hat{D} \in \mathbb{R}^{1 \times H \times W}$.
    \item \textbf{Normal Decoder:} Predicts a three-channel surface normal map $\hat{N} \in \mathbb{R}^{3 \times H \times W}$. A $L_2$ normalization layer is applied to the output to ensure the predicted vectors remain on the unit sphere.
    \item \textbf{Mask Decoder:} Predicts a single-channel contact mask $\hat{M} \in \mathbb{R}^{1 \times H \times W}$ representing the probability of physical contact at each pixel.
\end{itemize}

Each decoder branch mirror the encoder's depth using four upsampling blocks (\textit{Up}). We employ bilinear interpolation followed by a \textit{DoubleConv} block to reduce the channel dimension while concatenating skip connections from the corresponding encoder stages. Finally, a $1 \times 1$ convolution (\textit{OutConv}) is used to project the feature maps to the required output channels.

\subsubsection{Implementation Details}
The model is trained using the AdamW optimizer with a learning rate of $1 \times 10^{-3}$ and a weight decay of $1 \times 10^{-2}$. We employ a cosine annealing learning rate scheduler. In our implementation, the weights for the losses mentioned in Euqation~\eqref{unet_loss} are set to $\lambda_d=0.1, \lambda_n=1.0$, and $\lambda_m=5.0$.

\subsubsection{Tactile Perception Network Architecture}
The detailed architectural parameters of the Multi-Head U-Net are summarized in Table~\ref{tab:unet_architecture}. The encoder is shared across all tasks, while three independent decoder branches are utilized for depth, surface normal, and contact mask prediction.

\begin{table}[h]
\centering
\caption{Detailed architecture of the Multi-Task Tactile Perception U-Net. $C_{in}$ and $C_{out}$ denote the number of input and output channels, respectively.}
\label{tab:unet_architecture}
\small
\begin{tabular}{llcccl}
\hline
\textbf{Part} & \textbf{Layer} & \textbf{Type} & \textbf{$C_{in}$} & \textbf{$C_{out}$} & \textbf{Configuration} \\ \hline
\multirow{5}{*}{\rotatebox{90}{Encoder}} & \texttt{inc} & DoubleConv & 3 & 64 & Conv $3\times3$, Pad 1 \\
 & \texttt{down1} & Down & 64 & 128 & MaxPool $2\times2$ + DoubleConv \\
 & \texttt{down2} & Down & 128 & 256 & MaxPool $2\times2$ + DoubleConv \\
 & \texttt{down3} & Down & 256 & 512 & MaxPool $2\times2$ + DoubleConv \\
 & \texttt{down4} & Down & 512 & 512 & MaxPool $2\times2$ + DoubleConv \\ \hline
\multirow{5}{*}{\rotatebox{90}{Decoders\textsuperscript{*}}} & \texttt{up1} & Up & 1024 & 256 & Bilinear Up + Cat + DoubleConv \\
 & \texttt{up2} & Up & 512 & 128 & Bilinear Up + Cat + DoubleConv \\
 & \texttt{up3} & Up & 256 & 64 & Bilinear Up + Cat + DoubleConv \\
 & \texttt{up4} & Up & 128 & 64 & Bilinear Up + Cat + DoubleConv \\
 & \texttt{out} & OutConv & 64 & $N_{task}$ & Conv $1\times1$ \\ \hline
\end{tabular}
\vspace{1mm}
\begin{flushleft}
\footnotesize{\textsuperscript{*}Decoders for Depth ($N_{task}=1$), Normal ($N_{task}=3$), and Mask ($N_{task}=1$) share the same architecture but have independent weights.}
\end{flushleft}
\end{table}

\end{document}